\documentclass[iicol, sn-apa]{sn-jnl}

\usepackage{amsmath,amssymb}
\newlength\savewidth\newcommand\shline{\noalign{\global\savewidth\arrayrulewidth
  \global\arrayrulewidth 1pt}\hline\noalign{\global\arrayrulewidth\savewidth}}
  
\DeclareMathOperator{\avg}{avg}

\jyear{2023}%

\theoremstyle{thmstyleone}%
\theoremstyle{thmstyletwo}%

\theoremstyle{thmstylethree}%

\begin{document}

\title[Article Title]{S$^{2}$P$^{3}$: Self-Supervised Polarimetric Pose Prediction}

\author*[1]{\fnm{Patrick} \sur{Ruhkamp}}\email{p.ruhkamp@tum.de}
\equalcont{These authors contributed equally to this work.}

\author[1]{\fnm{Daoyi} \sur{Gao}}\email{daoyi.gao@tum.de}
\equalcont{These authors contributed equally to this work.}

\author[1]{\fnm{Nassir} \sur{Navab}}\email{n.navab@tum.de}
\author*[1]{\fnm{Benjamin} \sur{Busam}}\email{b.busam@tum.de}

\affil[1]{\orgdiv{TUM School of Computation, Information and Technology}, \orgname{Technical University of Munich}, \country{Germany}}

\abstract{This paper proposes the first self-supervised 6D object pose prediction from multimodal RGB+polarimetric images. The novel training paradigm comprises 1) a physical model to extract geometric information of polarized light, 2) a teacher-student knowledge distillation scheme and 3) a self-supervised loss formulation through differentiable rendering and an invertible physical constraint. Both networks leverage the physical properties of polarized light to learn robust geometric representations by encoding shape priors and polarization characteristics derived from our physical model. Geometric pseudo-labels from the teacher support the student network without the need for annotated real data. Dense appearance and geometric information of objects are obtained through a differentiable renderer with the predicted pose for self-supervised direct coupling. The student network additionally features our proposed invertible formulation of the physical shape priors that enables end-to-end self-supervised training through physical constraints of derived polarization characteristics compared against polarimetric input images. We specifically focus on photometrically challenging objects with texture-less or reflective surfaces and transparent materials for which the most prominent performance gain is reported.}




\maketitle

\section{Introduction}
\begin{figure*}[!hpt]
      \centering
      \includegraphics[width=1.0\textwidth]{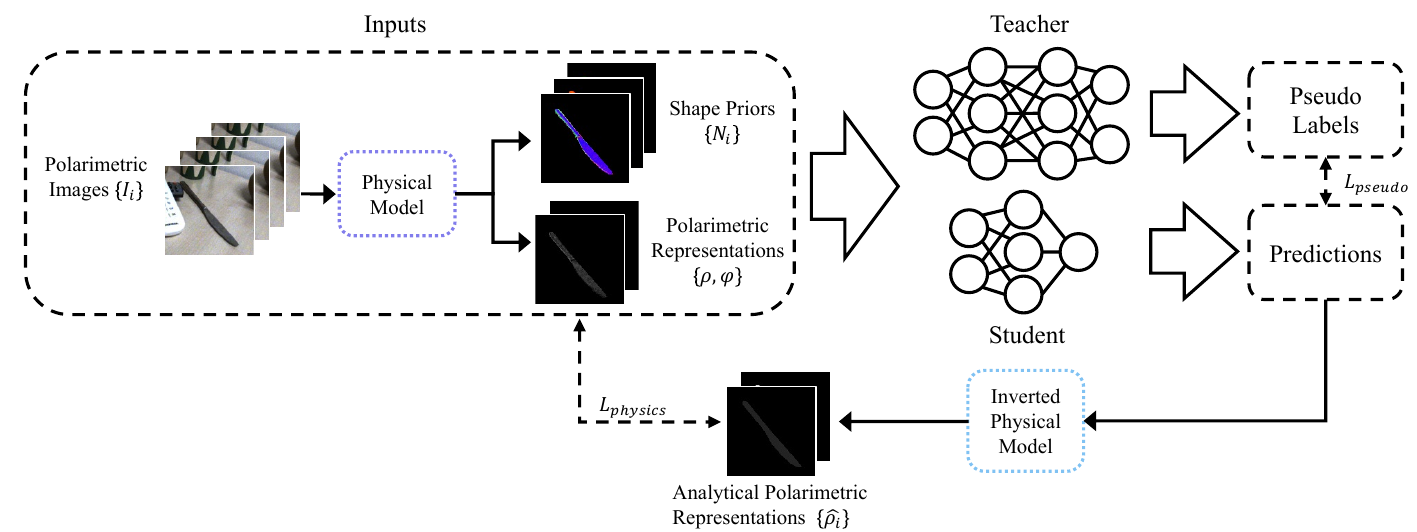}
      \caption{\textbf{\textbf{S}$^{2}$\textbf{P}$^{3}$ Pipeline Overview. } Our proposed teacher-student training scheme takes four polarization images taken under different polarization filter angles as well as polarimetric and geometrical representations derived from the analytical physical model as multi-modal inputs to both the teacher and student networks, individually. The student network is optimized not only towards the pseudo labels generated from the teacher denoted as $L_{pseudo}$, but also by $L_{physics}$ which minimizes the discrepancy between the polarimetric representations $\rho$ from the input images after the analytical physical model (cf. Inputs) and $\hat{\rho}$ derived through the inverted physical model from the predicted surface normal of the student network.
      }
      \label{fig:pipeline_small}
\end{figure*}
\label{intro}
"Fiat lux, et facta est lux".\footnote{Latin for "let there be light, and there was light".} 
Light has been the foundation of many significant scientific findings in history. Early horological devices utilized changing shadows cast from the sun to measure time throughout centuries across different civilizations all over the globe. Based on the constant speed of the electromagnetic wave (EM) with which light travels, it is possible to determine the distance of an object after emitting a light pulse by measuring its return time after reflection: a principle used in many active depth sensors. However, measurements are affected by artifacts such as multi-path interference (MPI)~\citep{cui20103d} due to reflective materials, ambient light~\citep{jung2021wild}, or inherently incorrect estimates when the light passes through transparent objects such as glass. This leads to inaccurate depth estimates, most noticeable for photometrically challenging objects~\citep{jung2022my}. Still, many methods that learn geometric tasks from images use such geometry information from depth data.

6D object pose estimation is one of those geometric tasks and essential in many computer vision and AR applications, ranging from robotics~\citep{wang2021demograsp} to safety-critical autonomous driving~\citep{Ost_2021_CVPR} and medical applications~\citep{busam2018markerless}. Recent methods integrate geometric information either directly as input~\citep{He_2021_CVPR} or leverage it for self-supervision~\citep{wang2021occlusion}. Reliable geometric cues can improve pose estimation performance, while unreliable and noisy depth information would interfere with what information a neural network has learned to extract.

Recent approaches integrate the geometric information of polarized light by learning features from both the estimated normal from polarization, and their polarization characteristics, for the task of 6D object pose estimation in a supervised way~\citep{gao2021polarimetric}. In the case of photometrically complex objects, it is shown that the deterioration of measured depth is even inferior to the use of this modality, ultimately making the direct geometric measurement obsolete.
The authors report impressive results for texture-less, reflective and translucent objects, outperforming state-of-the-art RGB-only~\citep{wang2021gdr} and RGB-D~\citep{He_2021_CVPR} methods. However, an extensive training dataset with ground-truth annotations is required, which may be challenging to obtain in practice, especially with high accuracy~\citep{PhoCal}. 

In \textbf{S}$^{2}$\textbf{P}$^{3}$, we study how a neural network can encode the geometric shape priors from polarized light captured with a multi-modal polarization camera for the task of 6D object pose estimation without the need for annotated real data. We leverage the aforementioned supervised polarimetric 6D object pose estimation method~\citep{gao2021polarimetric} as a teacher network and pre-train it on synthetically rendered polarimetric image data only. We then utilize its noisy predictions on real data, to support a student network with weak labels for guidance. A differentiable renderer is employed to enable self-supervision with dense geometric cues. Additionally, we propose an invertible formulation of the physical polarization model to analytically compute pixel-wise image characteristics from the geometric normal representation after the differentiable rendering of the student with the predicted 6D pose. This analytic inversion closes the self-supervision loop and allows for direct comparison with the input polarization as illustrated in Figure~\ref{fig:pipeline_small}. 

While we adopt the architecture of PPP-Net for a teacher network with an additional differentiable renderer, different from ~\citep{gao2021polarimetric}, we use this network to only train on synthetic data. This pre-trained model then produces predictions on the 6D pose of objects on real data, which are leveraged in our proposed teacher-student scheme as weak labels. The teacher network, based on PPP-Net, is thus merely one element of the overall method \textbf{S}$^{2}$\textbf{P}$^{3}$ as introduced here.

Inspired by the advancements in self-supervised learning and the use of differentiable renderers in end-to-end learning pipelines, as e.g. in Self6D++~\citep{wang2021occlusion}, we transfer such knowledge to the multi-modal imaging domain of polarization. Unlike Self6D++, where a renderer produces geometric information in terms of a depth map, which is then compared against a presumably noisy depth map from an active depth sensor, i.e., as explained in later sections here, we carefully study the physical properties of light and integrate encoded shape priors into a self-supervised scheme. This is possible through the differentiable analytical derivation of the physical properties from surface normal information.

The full pipeline of \textbf{S}$^{2}$\textbf{P}$^{3}$ thus includes a) novel architectural designs for the encoding of physical shape priors that extend the findings from PPP-Net to a student-teacher scheme; b) integrates RGB-agnostic shape information as surface normal maps from a differentiable renderer, offering a more resilient alternative to the issues posed by active depth sensors for photometrically complex objects; c) entails weak pseudo-labels in the form of geometric and pose information for self-supervision from the teacher network; and most notably, d) proposes an inverted physical model to leverage shape priors. 
The lightweight student network predicts and encodes these into a surface normal representation through a differentiable renderer. This encoded representation is then utilized to derive the object's analytical polarimetric representation. By integrating this representation into a new physical loss, we achieve complete end-to-end self-supervision using raw polarimetric images.

To this end we contribute in summary:

\begin{enumerate}
    \item \textbf{S}$^{2}$\textbf{P}$^{3}$ as a \textbf{hybrid neural-physics} approach to learn \textbf{6D object pose} prediction with photometric challenges through \textbf{self-supervision} with \textbf{neural encodings of geometric shape priors} from mutli-modal data.
    \item Insights on the interplay of \textbf{differentiable rendering} with the \textbf{invertible physical model} through extensive experiments on objects of \textbf{varying photometric complexity}.
    \item An \textbf{instance-level synthetic polarimetric image dataset} for 6D pose estimation that comprises objects present in PPP-Net~\citep{gao2021polarimetric} and PhoCal~\citep{PhoCal}.
\end{enumerate}

\section{Related Work}
\label{rw}
We revise related work in the realm of polarimetric imaging and 6D object pose estimation, including relevant datasets and recent self-supervised approaches, to provide a solid overview in the research field.

\paragraph{Polarimetric Imaging}
Early works on shape from polarization (SfP) investigate how the relation between polarization and the object's surface can be used to estimate surface normals and depth information, but focus on lab scenarios with controlled conditions of the environment~\citep{atkinson2006recovery,garcia2015surface,smith2018height,yu2017shape}.
These methods only rely on monocular polarization images, but multiple views can also be used for SfP~\citep{atkinson2005multi,cui2017polarimetric}, also extending to depth estimation~\citep{CroMo} from a freely moving camera. In~\citep{CroMo} the goal is to predict dense depth for outdoor scenes with photometrically easy objects in a (partly) supervised manner with depth measurements from an active structured light sensor while leveraging multi-modal input to account for other artefacts that affect depth predictions. 
Polarimetric images are also combined with photometric information from either stereo~\citep{atkinson2017polarisation} or monocular RGB~\citep{zhu2019depth} to complement each other for depth predictions. Polarized light can also improve initial noisy depth maps from other sensors~\citep{kadambi2017depth}. \citep{ba2020deep} compute a set of plausible cues from polarimetric images to predict surface normals with a neural network which can disambiguate such cues for SfP.
\citep{Lei_2022_CVPR} present a novel method for scene-level surface normal estimation from a single polarization image. By introducing a unique real-world dataset and employing advanced neural architecture with a multi-head self-attention module and viewing encoding, the study achieves superior performance in complex scenes.
Our approach is inspired from these findings to complement the pose estimation with shape priors from physical properties extracted from the polarized light. 

\paragraph{6D Object Pose Estimation}
Dense correspondence-based methods~\citep{zakharov2019dpod,hodan2020epos,li2019cdpn,park2019pix2pose,shugurov2021dpodv2} gained popularity in recent years for 6D object pose estimation. 
The key idea is to train a neural network to predict 2D-3D correspondences between each object pixel in the image and the 3D location of the corresponding point on the object's surface. Those correspondences are consecutively used either with PnP+RANSAC~\citep{lepetit2009epnp,fischler1981random}, the Umeyama algorithm~\citep{umeyama1991least}, or direct regression to compute the 6D object pose.
Hierarchical feature representations are proposed in ZebraPose~\citep{su2022zebrapose}, and also zero-shot methods are being investigated for the task of 6D pose estimation~\citep{shugurov2022osop}.
Many works on correspondence-based methods~\citep{zakharov2019dpod,shugurov2021dpodv2,park2019pix2pose,li2019cdpn,hodan2020epos} are limited by the computationally expensive post-processing for the RANSAC-based pose solver. GDR-Net~\citep{wang2021gdr} and its follower SO-Pose~\citep{di2021so} use learning-based MLP networks to directly predict the target pose from the predicted dense correspondences to improve the computing efficiency. 
In \textbf{S}$^{2}$\textbf{P}$^{3}$ we build upon these findings to directly regress the object pose. 

\paragraph{Geometric Depth Information} 
FFB6D~\citep{He_2021_CVPR} introduces a tight coupling strategy from cross-modal information exchanges with a keypoint extraction~\citep{He_2020_CVPR} that leverages geometry from depth.
Also other methods like Uni6D~\citep{jiang2022uni6d}, ESA6D~\citep{mo2022es6d}, FS6D~\citep{he2022fs6d} and DGECN~\citep{cao2022dgecn} include depth information into their prediction pipelines.
These approaches however, all critically depend on depth quality which suffers for photometrically challenging objects~\citep{gao2021polarimetric}. Geometric cues from polarization could mitigate such issues.

\paragraph{Self-Supervision} 
Self-supervised learning avoids the problem of lacking properly labeled data. In the realm of 6D pose estimation, differentiable rendering is being used to render synthetic images with a predicted pose to compare against input images~\citep{sock2020introducing}.
Self6D~\citep{wang2020self6d} proposes such approach, where a network is first trained on synthetic RGB data and then fine-tuned on real RGB-D data without pose annotations in a self-supervised manner.
They use depth data to align the visual and geometric cues which is the core part in the self-supervision stage. 
Building on top of Self6D, Self6D++~\citep{wang2021occlusion} replaces the one-stage pose regression backbone to two-stage GDR-net~\citep{wang2021gdr} backbone, and additionally introduces a pose refiner on top of the teacher network to improve the accuracy and the robustness towards occlusions. 

\paragraph{Polarimetric 6D Pose Prediction}
With recently published annotated datasets for real-world polarimetric category-level~\citep{PhoCal} and instance-level~\citep{gao2021polarimetric} 6D pose estimation, it is now possible to study methods with this mostly unexplored imaging modality~\citep{jung2023importance}.
PPP-Net~\citep{gao2021polarimetric} investigates the advantages of using polarization for supervised object pose estimation, and designs a hybrid pipeline leveraging polarization through a combination of physical model cues with learning, yielding impressive performance for photometrically challenging objects when compared against RGB and RGB-D baselines. 
However, acquiring real training data with accurate annotations is still difficult and not easily reproducible for other scholars without complex and expensive hardware~\citep{PhoCal,gao2021polarimetric}. 
Inspired by the strengths of polarimetric information in the supervised learning, we investigate the logical, yet non-trivial, next step towards exploring how this interesting modality can be integrated into a self-supervised scheme to reduce the need for annotated data. 
Different from Self6D~\citep{wang2020self6d} and Self6D++~\citep{wang2021occlusion}, we leverage polarimetric images, and extend the differentiable renderer to yield - besides appearance information - geometric representations in terms of normal maps of the object of interest. We further utilize this representation to compute polarimetric properties used for additional self-supervision through our proposed invertible physical model. To the best of our knowledge, we present the first method to utilize the geometric information from polarization in a self-supervised learning scheme.

\section{Polarimetric Physical Model}
\label{polarization}
\begin{figure}[!t]
      \centering
      \includegraphics[width=\linewidth]{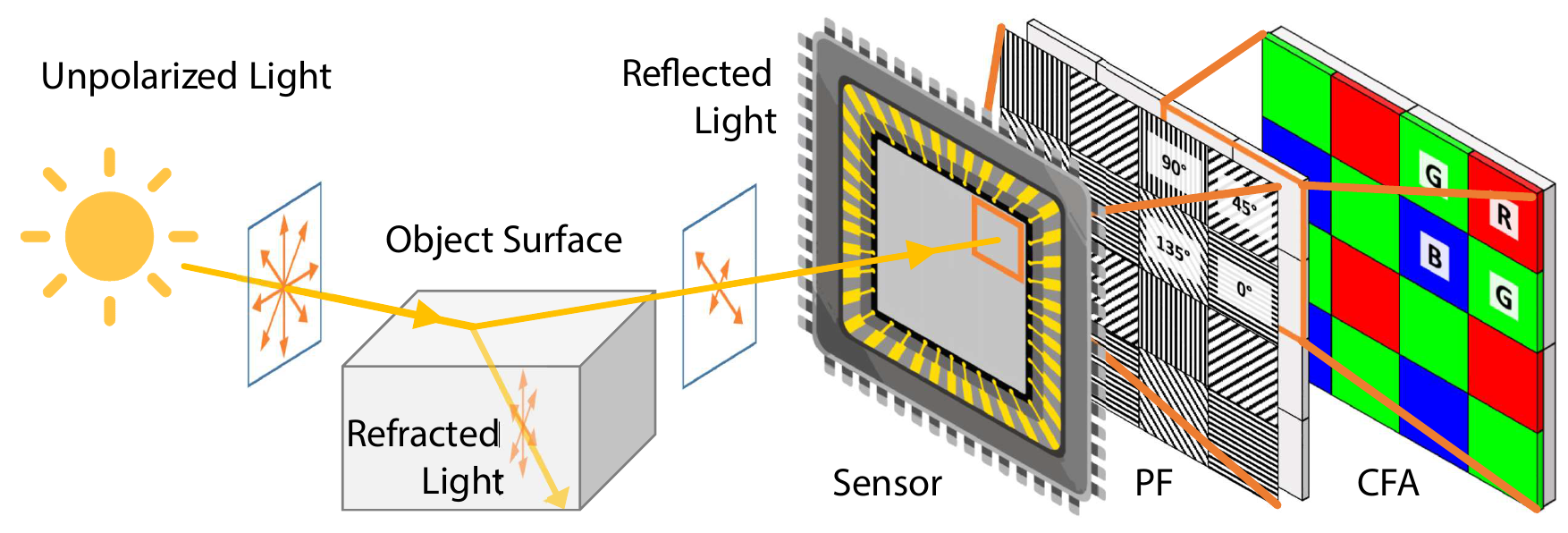}
      \caption{\textbf{Polarization Camera. }
      When an unpolarized light source reflects on an object surface, the resulting reflection comprises both a refracted and a reflected part, both of which are partially polarized. A polarization sensor captures this reflected light. In front of each pixel of the sensor, there are four polarization filters (PF) arranged at different angles: $0^{\circ}$, $45^{\circ}$, $90^{\circ}$, and $135^{\circ}$. Additionally, a colour filter array (CFA) is used to separate the reflected light into different wavebands.
      }
      \label{fig:rgbp_sensor}
\end{figure}

Commonly used sensors in computer vision send or receive light to measure the wavelength and energy within some specific spectrum. 
Additionally to this information, the relative oscillation of the electromagnetic wave defines its polarization. Emitted unpolarized natural light becomes polarized after being reflected from a surface, hence it carries information about the object's surface characteristics. 
The utilization of RGB-D sensors in pose estimation has gained popularity owing to their cost-effectiveness and easy integration into various devices. These sensors utilize active illumination for depth measurement, either through projection of a pattern or time-of-flight measurements. However, they are prone to photometric challenges such as translucency and reflections that can result in erroneous depth estimates. This paper presents a solution to these challenges through the use of surface normals derived from polarization of an RGB-P sensor (refer to Figure~\ref{fig:rgbp_sensor}).
After discussing some issus of RGB-D sensors, this section will introduce how aforementioned information can be measured with a passive sensor with integrated polarization filters. Then we will introduce how the physical model computes geometric shape priors from the information encoded in the polarimetric images and how our invertible formulation is integrated into our network architecture to enable direct self-supervision.

\subsection{Photometric Challenges for RGB-D}
\label{sec:rgbd}
\begin{figure}[!t]
      \centering
      \includegraphics[width=\columnwidth]{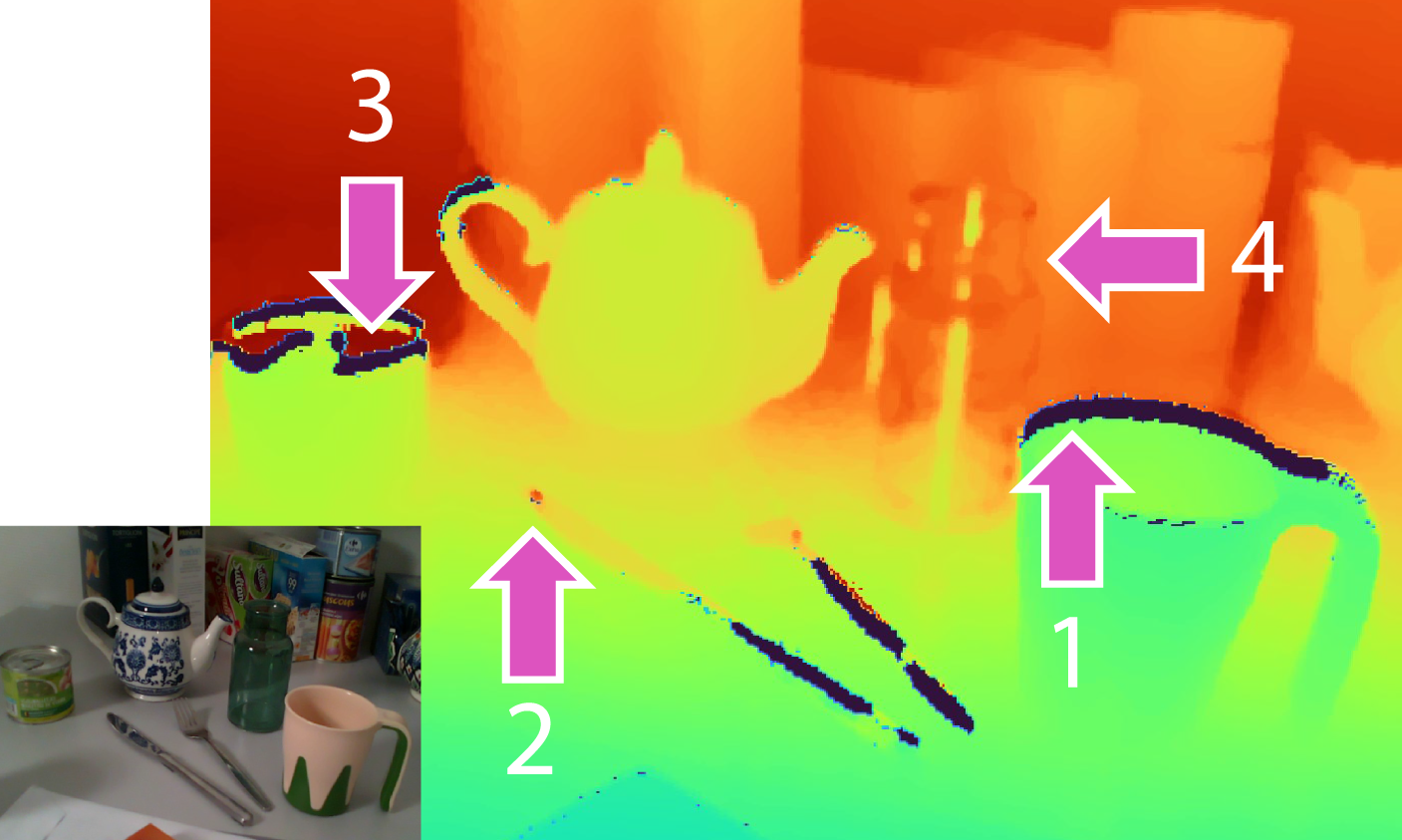}
      \caption{\textbf{Depth Artifacts. }
      The RealSense L515 depth sensor exhibits miscalculations in depth values for common household objects. Specifically, boundaries (1,3) invalidate pixels, and strong reflections (2,3) lead to incorrect depth estimates that are too far from the true value. In the case of semi-transparent objects like the vase (4), the depth sensor has difficulty detecting them, resulting in partially invisible objects and inaccurate measurements of the distance to objects behind them.
      }
      \label{fig:rgbd_issues}
\end{figure}

Commercial depth sensors rely on photometric measurements to estimate depth, by using active illumination either by projecting a pattern (e.g. intel RealSense D series) or using time-of-flight (ToF) measurements (e.g. Kinect v2 / Azure Kinect, intel RealSense L series). 
This makes them susceptible to challenges such as reflections and translucency, which can artificially extend the roundtrip time of photons or deteriorate the projected pattern. 
As a result, accurate depth estimation becomes infeasible in such scenarios, as illustrated in Figure~\ref{fig:rgbd_issues} for a set of common household objects.
The ToF sensor (RealSense L515) used in the experiment struggles to detect the semi-transparent vase, which appears almost invisible to the sensor. Additionally, reflections on the \textit{cutlery} and \textit{can}, cause the sensor to generate depth estimates that are significantly further from the true value, while strong reflections at boundaries result in pixel distances that are invalidated.

\subsection{Surface Normals from Polarization}
\label{sec:polarization}
\begin{figure}[!t]
      \centering
      \includegraphics[width=\columnwidth]{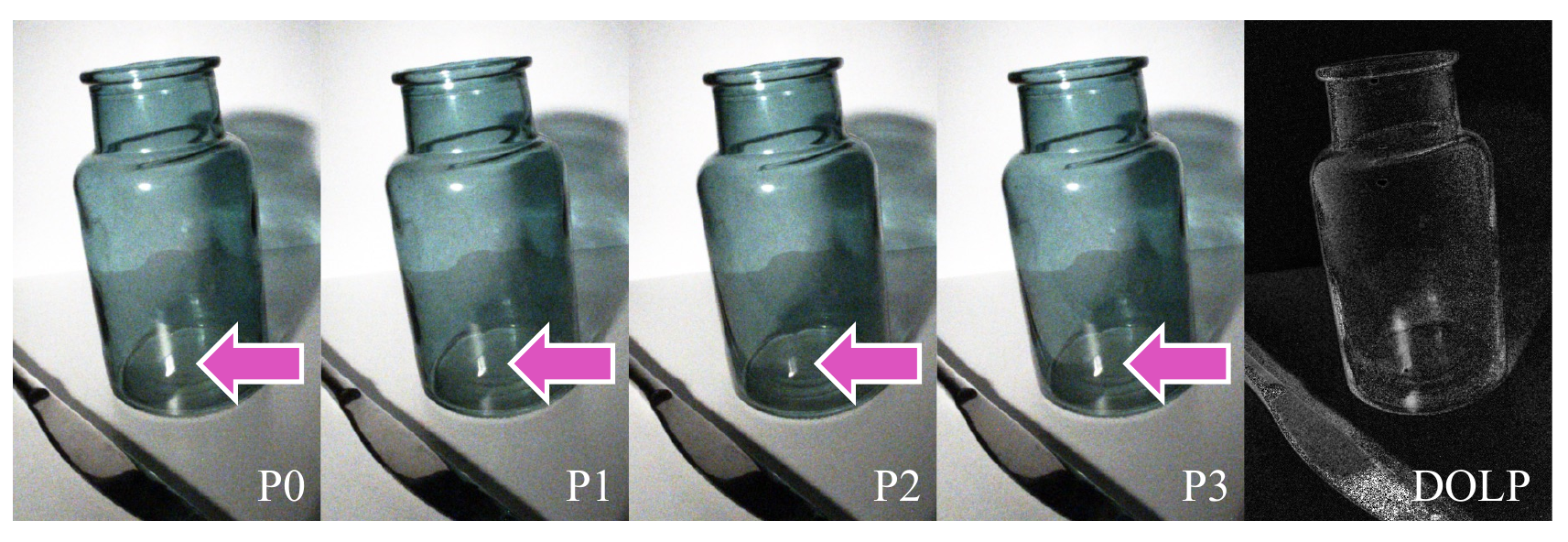}
      \caption{\textbf{Degree of Polarization. }
      The polarization of light changes when it reflects off a translucent surface, resulting in differences in the polarimetric image quadruplet, with different polarization angles (P0-P3), that are directly related to the surface normal. In particular, the degree of polarization (DoP) for both the translucent and reflective surfaces is considerably higher than for the rest of the image, as shown in the indicated areas in the image.
      }
      \label{fig:pol_example}
\end{figure}

Most artificial and natural light is unpolarized, meaning the electromagnetic wave oscillates along all planes perpendicular to the direction of propagation of the light~\citep{fliessbach2012elektrodynamik}.
When unpolarized light passes through a linear polarizer or is reflected at Brewster's angle from a surface, it becomes perfectly polarized. The refractive index of a material determines how fast light travels through it, how much of it is reflected, and the Brewster's angle of that medium. When light is reflected at the same angle to the surface normal as the incident ray, we call it specular reflection. The remaining part penetrates the object as refracted light, which becomes partially polarized as it traverses through the medium. This light wave escapes from the object and creates diffuse reflection~\citep{fliessbach2012elektrodynamik}. We use Figure~\ref{fig:pol_example} to provide an example that illustrates these concepts.

For real physical objects, the resulting reflection is a combination of specular and diffuse reflection, where the ratio largely depends on the refractive index and the angle of incident light. We propose to use surface normals obtained from polarization to overcome the photometric challenges faced by RGB-D sensors. Our method can be applied to various applications, including pose estimation, where accurate 3D information is crucial. 

\subsection{Image Formation Model}
We present the fundamental polarization image formation model and our invertible physical model that links the polarimetric and geometrical representations. 
When light with a specific intensity $I$ and wavelength $\lambda$ reaches the sensor, it passes through the color filter array (CFA), which separates the light into RGB wavebands, as shown in Figure~\ref{fig:rgbp_sensor}. 
The incoming light also has a degree of polarization (DoP) $\rho$ and a direction (angle) of polarization (AoP) $\phi$.
As light passes through a polarizer array on top of a pixel unit with four different polarization angles $\varphi_{pol} \in \{0^\circ, 45^\circ, 90^\circ, 135^\circ\}$, the oscillation state of light is recorded alongside its wavelength and energy~\citep{kalra2020deep}. 
The polarization image formation model in Equation~\ref{eq:i_pol} defines the underlying parameters that contribute to the captured polarized intensities as:
\begin{align}
    \label{eq:i_pol}
    I_{\varphi_{pol}} &= I_{un} \cdot \ (1+\rho \  \cos(2(\phi - \varphi_{pol}))),
\end{align}
where the unpolarized intensity $I_{un}$ can be computed via averaging over polarized intensities $I_{\varphi_{pol}}$ under different polarization filter angles $\varphi_{pol} \in \{0^{\circ}, 45^{\circ}, 90^{\circ}, 135^{\circ}\}$. 
The degree of polarization (DoP) $\rho$ and angle of polarization (AoP) $\phi$ can be solved from a linear least squares system~\citep{huynh2010shape} from a set of polarization images captured under different polarization filter angles as:
\begin{equation}
\label{eq:i_overdetermin}
\begin{aligned}
   \begin{bmatrix}
        I_{\varphi_{pol, 1}}\\
        \vdots \\
        I_{\varphi_{pol, 4}}\\
    \end{bmatrix}
    = 
    \begin{bmatrix}
        1 & \cos{2\varphi_{pol, 1}} & \sin{2\varphi_{pol, 1}}\\
        & \vdots  &  \\
        1 & \cos{2\varphi_{pol, 4}} & \sin{2\varphi_{pol, 4}}\\
    \end{bmatrix}
    \begin{bmatrix}
        x_1 \\
        x_2 \\
        x_3 \\
    \end{bmatrix},
\end{aligned}
\end{equation}
where the unknowns $x_i$ in the linear system represent $x_1=I_{un}$, $x_2=I_{un}\rho\cos{2\phi}$, and $x_3=I_{un}\rho\sin{2\phi}$.

We find $\varphi$ and $\rho$ from the over-determined system of linear equations in \ref{eq:i_pol} using linear least squares. Depending on the surface properties, AoP is calculated as:
\begin{align}
    \left\{
    \begin{array}{lll}
        \phi_{d} [\pi] &= \alpha \ &\text{for diffuse reflection}\\
        \phi_{s} [\pi] &= \alpha - \frac{\pi}{2} \ &\text{for specular reflection}
    \end{array}
    \right.
    ,
    \label{eqn:aolp}
\end{align}
where $[\pi]$ indicates the $\pi$-ambiguity and $\alpha$ is the azimuth angle of the surface normal $\textbf{n}$.
We can further relate the viewing angle $\theta \in [0, \pi/2]$ to the degree of polarization by considering Fresnel coefficients, thus DoP is similarly given by~\citep{atkinson2006recovery}:
\begin{align}
    \left\{
    \begin{array}{l}
        \rho_{d} = \frac
            {(\eta-1/\eta)^{2}\sin^{2}(\theta)}
            {2+2\eta^{2}-(\eta+1/\eta)^{2}\sin^{2}(\theta)+4\cos(\theta)\sqrt{\eta^{2}-\sin^{2}(\theta)}}
            \\
            \\
        \rho_{s} =  \frac
            {2\sin^{2}(\theta)\cos(\theta)\sqrt{\eta^{2}-\sin^{2}(\theta)}}
            {\eta^{2}-\sin^{2}(\theta)- \eta^{2}\sin^{2}(\theta) +2\sin^{4}(\theta)}
    \end{array}
    \right.
    ,
    \label{eqn:dolp}
\end{align}
with the refractive index of the observed object material $\eta$.
Solving equation~\ref{eqn:dolp} for $\theta$, we retrieve three solutions $\theta_d,\theta_{s1},\theta_{s2}$, one for the diffuse case and two for the specular case. For each of the cases, we can now find the 3D orientation of the surface by calculating the surface normals:
\begin{equation}
    \label{eq:normals}
    \mathbf{n} = \left(
    \cos{\alpha}\sin{\theta}, 
    \sin{\alpha}\sin{\theta},
    \cos{\theta}
    \right)^{\text{T}}.
\end{equation}

We use these plausible normals $\mathbf{n}_{d}, \mathbf{n}_{s1}, \mathbf{n}_{s2}$ as physical priors per pixel as input to the neural network.

With the help of the physical model defined by Equations~\ref{eq:i_pol} and~\ref{eq:i_overdetermin}, we can now derive physical polarimetric characteristics which encode shape information as geometric normals. 
More formally, when light gets reflected by the object's surface, the shape information is encoded in the captured polarization intensities accordingly.
The physical model in our pipeline reveals the implicitly encoded shape information to provide object-centric priors orthogonal to intensity information. We derive a set of explicit object shape priors $N_{i}$ based on polarimetric intensities $I_{\varphi_{pol}}$ and properties $\rho, \phi$ as~\citep{ba2020deep, zou20203d}. The ambiguities within this process lead to non-unique solutions as in~\citep{ba2020deep}, yet we encode them in a pixel-exclusive manner to guide the network to distinguish between different priors and extract meaningful geometrical features.

\subsection{Invertible Physical Model}
Inverting the model and assuming a given normal map of an object, e.g., from a differentiable renderer with an estimated 6D pose as in our training scheme, we define an invertible solution to solve for the polarimetric representation analytically. 
This serves to close the loop from the network's prediction by transferring the information of the object's pose parameterized as 6D transformation through a differentiable renderer into a geometric form and further into encoded physical properties of light reflections that can be compared against the original input information in a self-supervised scheme.

The inverted physical model aims to bring a loop closure from the other end by taking the rendered object surface normal map to analytical polarimetric parameters considering different reflection properties. 
We obtain the viewing angle $\theta_v$ from $\cos{\theta_v} = \mathbf{n}	\cdot \mathbf{v}$ where $\mathbf{n}$ is the rendered object surface normal map, and the viewing vector $\mathbf{v}$ is defined as $\mathbf{v} = -\pi^{-1}(u,v,K)$ with $\pi^{-1}$ which serves as backprojection operation for pixel $(u,v)$ with camera intrinsics $K$. The analytical DoP $\hat{\rho}$ is then derived via formulations for diffuse and specular reflection cases:
\begin{align}
    \left\{
    \begin{array}{l}
        \hat{\rho_{d}} = \frac
            {(\eta-1/\eta)^{2}\sin^{2}(\theta_v)}
            {2+2\eta^{2}-(\eta+1/\eta)^{2}\sin^{2}(\theta_v)+4\cos(\theta_v)\sqrt{\eta^{2}-\sin^{2}(\theta_v)}}
            \\
            \\
        \hat{\rho_{s}} =  \frac
            {2\sin^{2}(\theta_v)\cos(\theta_v)\sqrt{\eta^{2}-\sin^{2}(\theta_v)}}
            {\eta^{2}-\sin^{2}(\theta_v)- \eta^{2}\sin^{2}(\theta_v) +2\sin^{4}(\theta_v)}
    \end{array}
    \right.
    ,
    \label{eqn:analytical_dolp}
\end{align}
where $\eta$ is a constant defined by the refractive index of object materials. The inverted physical model offers the possibility to optimize the model via object shape cues, which is more robust in photometrically challenging scenarios compared to active depth sensors.

\section{Methodology}
\label{methods}

\begin{figure*}[!t]
      \centering
      \includegraphics[width=1.0\textwidth]{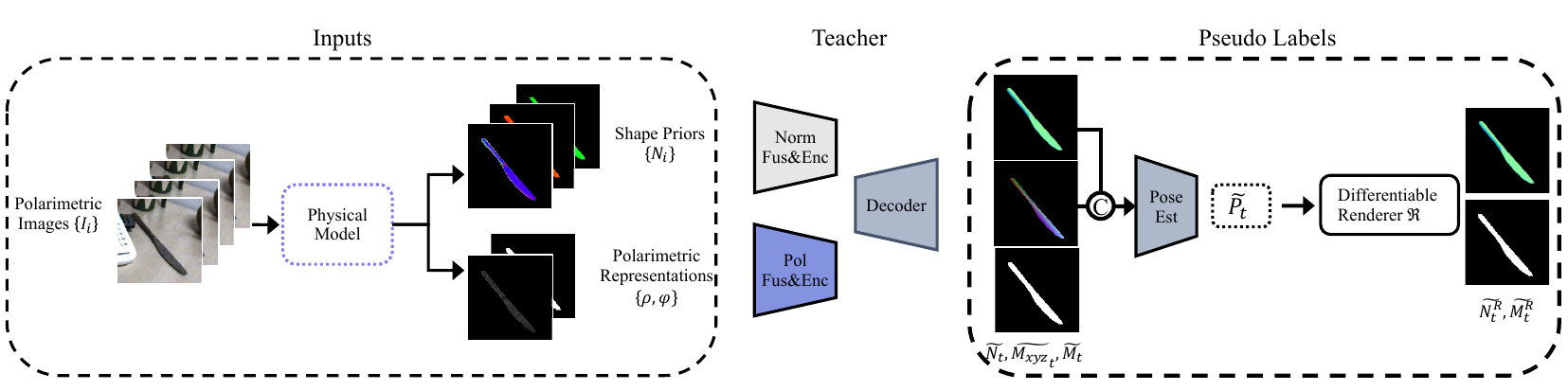}
      \caption{\textbf{\textbf{S}$^{2}$\textbf{P}$^{3}$ Teacher Network. } The network takes the shape priors and polarimetric representations, both derived from the analytical physical model from four polarized images, as input. Before retrieving the 6D object pose, intermediate geometrical representations are predicted. A differentiable renderer utilizes the predicted pose to provide a rendered normal map and object mask.}
      \label{fig:pipeline_teacher}
\end{figure*}

The objective of \textbf{S}$^{2}$\textbf{P}$^{3}$ is to achieve 6D object pose prediction without relying on annotated real data. To accomplish this, a teacher-student training approach is suggested, which utilizes pre-training on synthetic data and pseudo-labels from the teacher during self-supervision as depicted in Figure~\ref{fig:pipeline_small}. By additionally incorporating the proposed invertible physical model for self-supervision, \textbf{S}$^{2}$\textbf{P}$^{3}$ makes full use of the geometric data encoded in the polarimetric images. This section outlines the hybrid polarization-based pipeline for learning object pose and explains the physics-induced self-supervision approach in detail.

\subsection{\textbf{S}$^{2}$\textbf{P}$^{3}$ Network Architecture}
\label{sec:network}

\begin{figure*}[!hpt]
      \centering
      \includegraphics[width=1.0\textwidth]{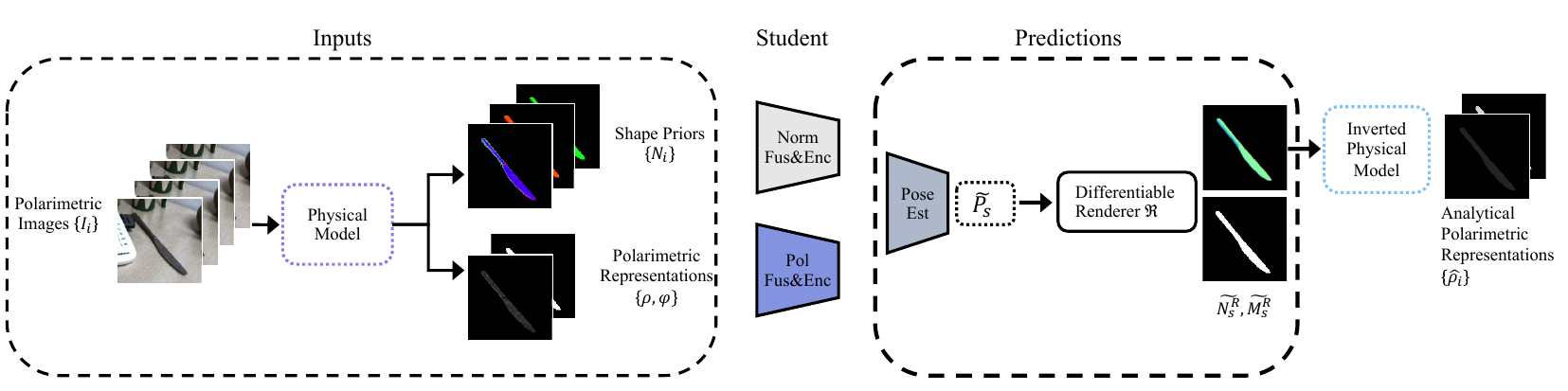}
      \caption{\textbf{\textbf{S}$^{2}$\textbf{P}$^{3}$ Student Network. } Different from the teacher network in Figure~\ref{fig:pipeline_teacher}, the student is more light-weight by neglecting the explicit decoding of predicted geometric representations.
	  }
      \label{fig:pipeline_student}
\end{figure*}

\begin{figure*}[!hpt]
      \centering
      \includegraphics[width=1.0\textwidth]{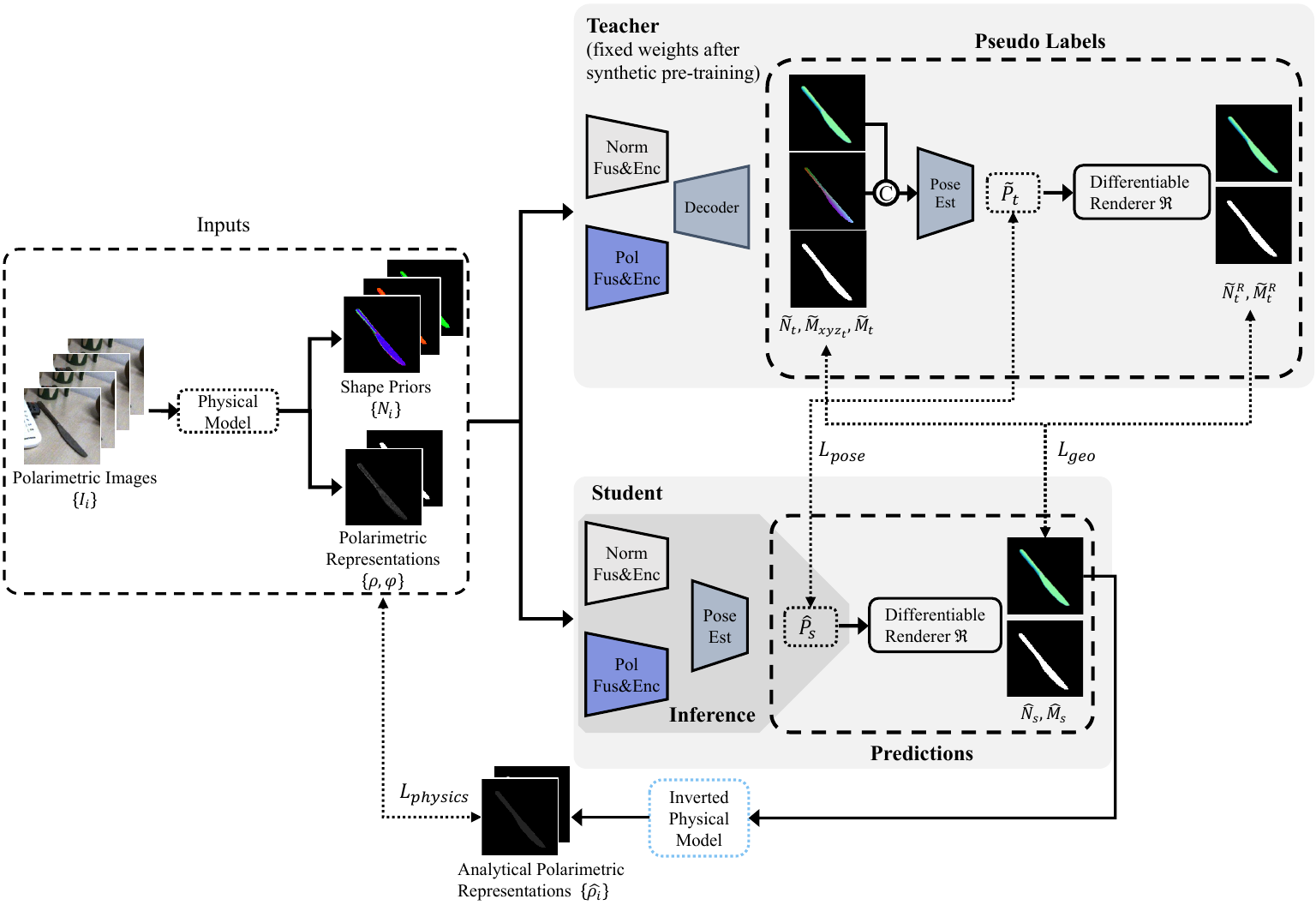}
      \caption{\textbf{\textbf{S}$^{2}$\textbf{P}$^{3}$ Pipeline Overview. } Our proposed teacher-student training scheme takes four polarization images taken under different polarization filter angles as well as polarimetric and geometrical representations derived from the physical model as inputs to both the teacher and student networks. The student network is optimized not only towards the pseudo labels generated from the teacher denoted as $L_{pseudo}$, but also by $L_{physics}$ which minimizes the discrepancy between $\rho$ from the physical model and $\hat{\rho}$ from the inverted physical model. During inference, the lightweight student network only predicts direct pose estimates as indicated by the gray background color.
      }
      \label{fig:pipeline}
\end{figure*}

\textbf{S}$^{2}$\textbf{P}$^{3}$, consisting of a teacher network (cf. Figure~\ref{fig:pipeline_teacher}) with a larger capacity and a light student network (cf. Figure~\ref{fig:pipeline_student}), is illustrated in Figure~\ref{fig:pipeline} as a schematic overview. Both networks are pre-trained on synthetic data, whereas the teacher later provides pseudo labels on real data to guide the student network in a self-supervised manner. 
The detailed architecture illustrates essential extensions, modifications, and important design choices of \textbf{S}$^{2}$\textbf{P}$^{3}$ compared against established student-teacher training schemes in the community of 6D object pose estimation~\citep{wang2021occlusion}. These are explained in detail in the following and justified with ablations in our experiments section. 

\paragraph{Teacher Network}
Inspired by the architecture of PPP-Net~\citep{gao2021polarimetric}, we propose our polarimetric network with an extended differentiable renderer, as the teacher of \textbf{S}$^{2}$\textbf{P}$^{3}$ (cf. Figure~\ref{fig:pipeline_teacher}). Here, the inputs of polarimetric intensities and geometrical shape priors are encoded through separate input heads, followed by an explicit decoder to predict an object mask $\mathbf{\Tilde{M}_t}$, an object normal map $\mathbf{\Tilde{N}_t}$, and the dense correspondences as normalized object coordinate map $\mathbf{\Tilde{M}_{{xyz}_t}}$. 
The spatial and shape correlation of $\mathbf{\Tilde{M}_{xyz_t}}$ and $\mathbf{\Tilde{N}_t}$ serve as inputs to an object pose estimation module~\citep{wang2021gdr}, in which the predicted rotation vector is parameterized in the form of allocentric continuous 6D representation~\citep{zhou2019continuity} and the predicted translation as scale-invariant vector~\citep{li2019cdpn}. 
We further convert them into a standard rotation matrix $\mathbf{\Tilde{R}_t} \in \mathbb{R}^{3 \times 3}$ 
and a translation vector $\mathbf{\Tilde{t}_t} \in \mathbb{R}^3$ 
and denote the final pose as 
$\mathbf{\Tilde{P}_t} = [\mathbf{\Tilde{R}_t} \mid \mathbf{\Tilde{t}_t}] $. 
Here, we extend the neural network of PPP-Net.
To compute pixel-wise geometrical pseudo labels from the predicted pose, a differentiable renderer takes the object's CAD model and $\mathbf{\Tilde{P}_t}$ as inputs to render an object mask $\mathbf{\Tilde{M}^R_t}$ and an object normal map $\mathbf{\Tilde{N}^R_t}$. 
All the predicted and rendered quantities serve as weak pseudo labels for the student network.

\paragraph{Student Network. }
We propose a lightweight student network without explicit geometric decoder, different to Self6D++~\citep{wang2021occlusion}, where the network directly regresses the predicted pose for the student $\mathbf{\hat{P}_s}$ (cf. Figure~\ref{fig:pipeline_student}). 
This also favors fast inference while maintaining high accuracy. 
Our ablations, discussed later in Table~\ref{tab:ablation_architecture}, indicate the superiority of our student network design. 
The teacher network consists of about 5.5 million weights, whereas our lightweight teacher does not need the explicit decoder, thus reducing the network to about 5 million weights. While the number of parameters is not significantly reduced, the inference time and also pose prediction accuracy is greatly improved by not predicting the intermediate geometric representations, as discussed later in the results section.
We test this against the design choice of Self6D++~\citep{wang2021occlusion} of having the student network identical to the teacher but without a subsequent pose refiner. 
Our student network converges towards better predictions without the redundant explicit prediction of intermediate geometric representations with our proposed self-supervision. 
The final output of our student in \textbf{S}$^{2}$\textbf{P}$^{3}$, thus only consists of the predicted pose $\mathbf{\hat{P}_s}$. 
To link the predictions with geometric and polarimetric properties, we render an object normal map $\mathbf{\hat{N}_s}$ and an object mask $\mathbf{\hat{M}_s}$ given $\mathbf{\hat{P}_s}$ via the differentiable renderer - analogous to the teacher network. 
We will detail how this polarimetric representation of the geometric information is utilized in a self-supervised loss term in the following.

\subsection{Physics-Induced Self-Supervised Training Scheme}

As detailed before, the polarimetric images contain rich information that we provide as explicit representations to the network to learn neural geometric encodings. This section defines how these representations are further leveraged and integrated into our physically induced self-supervised scheme, firstly through implicit and explicit weak pseudo-labels of the teacher network, and second as direct coupling by closing the loop towards the input information of the pipeline.

\label{sec:loss_formulations}
\subsection{Loss Formulations}
Our proposed optimization scheme comprises two complementary paradigms. 
The first passes knowledge of the pre-trained teacher to the student in the form of weak labels of the pose $\mathbf{\Tilde{P}_t}$ and related object shape knowledge $\{\mathbf{\Tilde{M}_t}, \mathbf{\Tilde{N}_t}, \mathbf{\Tilde{M}^R_t}, \mathbf{\Tilde{N}^R_t}\}$, which we define as pseudo label loss $\mathcal{L}_{pseudo}$. 
The second is to utilize the inverted physical model to optimize the student prediction $\mathbf{\hat{P}_s}$ via raw polarization data in our physical loss term $\mathcal{L}_{physics}$ detailed below. 

To account for potential misalignment between the decoded shape knowledge $\{\mathbf{\Tilde{M}_t}, \mathbf{\Tilde{N}_t}\}$ and pose knowledge $\mathbf{\Tilde{P}_t}$, we compare the predicted mask $\mathbf{\Tilde{M}_t}$ and the rendered mask $\mathbf{\Tilde{M}^R_t}$ and normalize the discrepancy to a scalar value of $\delta$, which serves as the criteria of choosing pseudo ground truth for the geometrical regularization term $\mathcal{L}_{geo}$ and a dynamic weighting term in the overall learning objective. The final formulation is then:
\begin{align}
\mathcal{L}_{pseudo} = \lambda_1\mathcal{L}_{pose} + \mathcal{L}_{geo},
\end{align}
with:
\begin{align}
    \label{eq:loss_pose_geo}
    \mathcal{L}_{pose} &= \underset{\mathbf{x} \in \mathcal{M}}{\avg} \| ({\mathbf{\Tilde{R}_t}} \mathbf{x} +\mathbf{\Tilde{t}_t}) - (\mathbf{\hat{R}_s} \mathbf{x} + \mathbf{\hat{t}_s}) \|_1, \\ 
    \mathcal{L}_{geo} &= \mathcal{L}_{mask} + \mathcal{L}_{normals},
\end{align}
in which we define the $\mathcal{L}_{mask}$ as mean squared error and $\mathcal{L}_{normals}$ as cosine similarity loss. The rendered representations $\{\mathbf{\Tilde{M}^R_t}, \mathbf{\Tilde{N}^R_t}\}$ are chosen as geometrical pseudo ground truth if $\delta$ is within a predefined threshold $r$, otherwise the predicted representations are selected, also leading to a reduced weighting factor $\lambda_1 = (1 - \delta)$ on direct pseudo pose loss $\mathcal{L}_{pose}$.

\paragraph{Physical Constraints}
To enable self-supervision via the invertible physical model, the rendered geometric normal map $\mathbf{\hat{N}_s}$ serves as input to solve for analytical diffuse and specular DoP $\{ \hat{\rho_{d}},\hat{\rho_{s}} \}$ according to Equation~\ref{eqn:analytical_dolp}. 
To benefit from the underlying physical process of polarimetric imaging, $\mathcal{L}_{physics}$ deploys a pixel-wise minimum selection mechanism inspired by~\citep{CroMo}:
\begin{equation}
    \label{eq:loss_physics}
    \mathcal{L}_{physics} = \underset{\mathbf{x} \in \{ \hat{\rho_{d}},\hat{\rho_{s}} \} }{\min} \| \rho - \mathbf{x} \|_1.
\end{equation}

To avoid the domain gap between the analytically solved intensity map and the real polarimetric images as in~\citep{CroMo}, we directly formulate the loss function based on polarimetric properties instead of polarimetric intensities. 
Hence, the student's output is optimized to align with raw DoP $\rho$ from real polarization images.

The overall loss combines the knowledge from the teacher and the raw data as:
\begin{align}
\mathcal{L} = \mathcal{L}_{pseudo} + \mathcal{L}_{physics}.
\end{align}

\section{Experimental Results}
\label{experiments}
We perform extensive evaluations and ablations on the instance-level polarimetric 6D pose dataset on which PPP-Net~\citep{gao2021polarimetric} provides a strong baseline against RGB-only~\citep{wang2021gdr} and RGB-D~\citep{He_2021_CVPR} state-of-the-art supervised methods~\citep{PhoCal}. This section first states implementation parameters for training, outlines the synthetic dataset generation, and describes the real polarimetric dataset. Detailed quantitative results on real data are discussed, and extensive ablations on different loss terms and modalities are analyzed. Our experiments specifically study the influence of polarimetric physical cues in a self-supervised scheme on objects of varying photometric complexity for instance-level 6D object pose prediction. Polarimetric images and self-supervised schemes are both mostly unexplored tracks in 6D pose estimation. As such, we take the supervised PPP-Net~\citep{gao2021polarimetric}, and the self-supervised Self6D++~\citep{wang2021occlusion} trained on RGB and RGB-D data, as strong baselines for comparison. Self6D++~\citep{wang2021occlusion} is the SOTA method in self-supervised 6D object pose estimation with RGB-D information, outperforming other baselines by a large margin~\citep{sock2020introducing,wang2020self6d}. As such, it represents a valid comparison and justifies the improvements of our method.
Likewise, PPP-Net~\citep{gao2021polarimetric} outperforms state-of-the-art RGB-only methods on photometrically challenging objects as under consideration here. Hence, it constitutes a legitimate representative of RGB-only methods as strong baseline for the experiments under consideration here.

\begin{figure*}[!t]
      \centering
      \includegraphics[width=\textwidth]{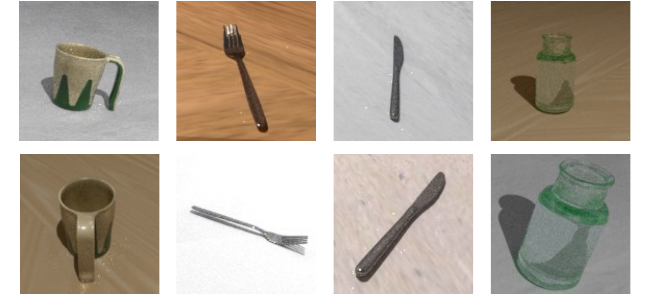}
      \caption{\textbf{Synthetic Dataset. }Samples of objects with varying photometric complexity are illustrated from different viewpoints.
      }
      \label{fig:syn_data}
\end{figure*}

\begin{figure*}[!t]
      \centering
      \includegraphics[width=\textwidth]{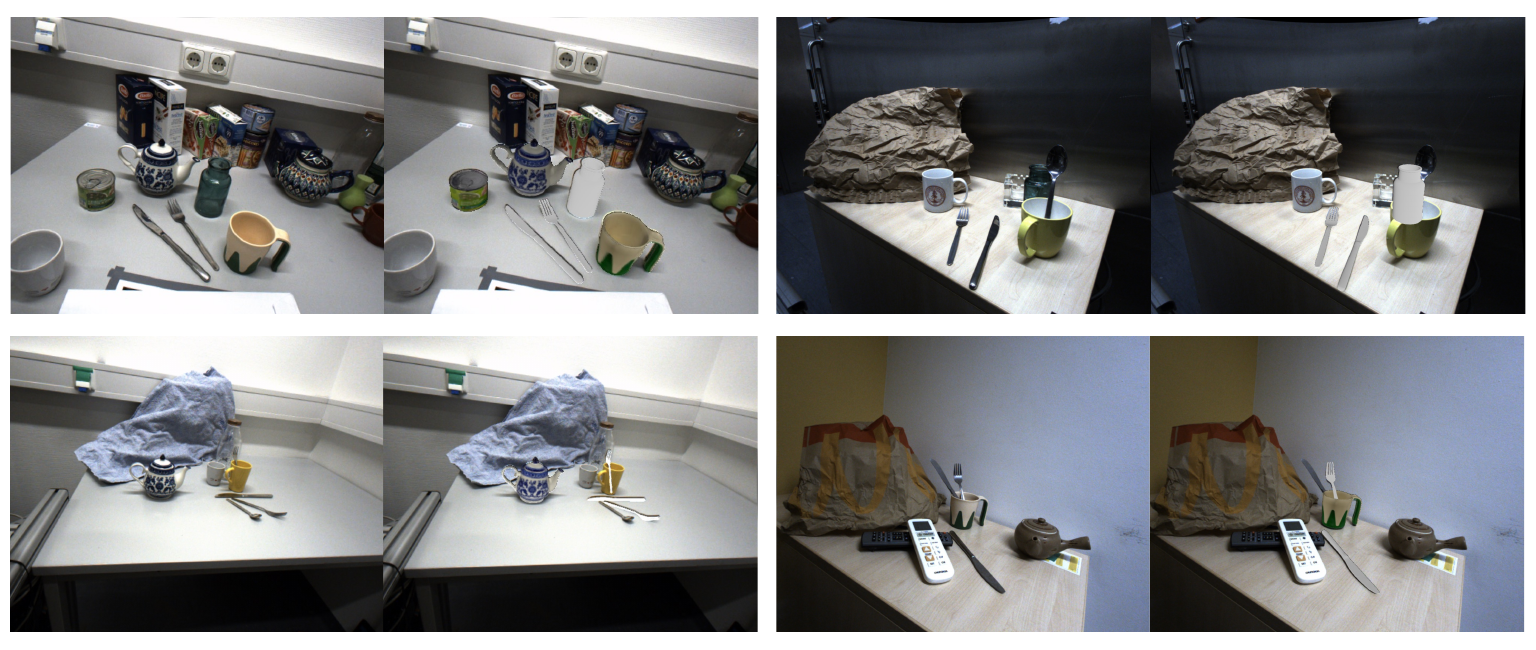}
      \caption{\textbf{Real Dataset. }Samples of objects are illustrated from different viewpoints. The rendered objects using GT pose illustrate the white-color rendered texture.
      }
      \label{fig:real_data}
\end{figure*}

\subsection{Synthetic Data Generation}

Given a CAD model of an object, we randomly sample camera locations on its upper hemisphere for rendering. To further enforce realistic renderings and to reduce the domain gap, we set up backgrounds with different textures and lighting positions in Mitsuba2 renderer~\citep{nimier2019mitsuba} to acquire 200-800 sets of polarization images for each object.

We present illustrations of our synthetic dataset for different viewpoints in Figure~\ref{fig:syn_data} to illustrate the variety of sampled poses, objects of different photometric complexity, and their appearance in the image. The synthetic dataset is used to pretrain the teacher and student networks. We render a set of four polarimetric images with different angles of the polarization filter according to the camera used in the real setup.

As rendering is very time-consuming, we provide the dataset~\footnote{\url{https://daoyig.github.io/}}.
We also train a customized object detector on synthetic data to later provide predicted masks and bounding boxes on the real domain.

We present samples of our real polarimetric dataset with annotated object poses in Figure~\ref{fig:real_data}. The objects rendered using ground truth pose labels indicate the high quality of data annotation, and the object models with white color rendering indicate their textureless nature, which supports our design of removing the need for color texture supervision.

\subsection{\textbf{S}$^{2}$\textbf{P}$^{3}$ Training}

We detail the two phases of the training: "Synthetic Pre-Training" on rendered data and "Self-supervised Training on Real Data." The former uses synthetic data with 6D pose annotations for supervised pre-training of the teacher and the student networks individually. In the latter phase, we use real data to train the student network in a self-supervised fashion by leveraging our proposed novel training scheme and loss function.

\subsubsection{Synthetic Pre-Training}
Both the teacher and student models go through a pre-training phase in which they receive supervision exclusively based on the 6D pose information, derived from ground truth annotations from synthetic data. During this phase, the loss function has a two-part structure: an L1 loss is utilized for translation, while a point matching loss is applied for rotation. Notably, the differentiable renderer is not integrated into this pre-training stage. In terms of computational time, the pre-training process takes several hours, typically ranging from 4 to 5 hours for each object. Subsequently, the self-supervised phase is more time-intensive, demanding approximately 10 hours per object.

\subsubsection{Self-supervised Training on Real Data} 
We evaluate our method on a specific data split of the instance-level 6D pose estimation dataset introduced in~\citep{gao2021polarimetric} containing objects with varying photometric complexity with highly accurate annotations from robotic forward-kinematics. The RGB-P data is acquired with the polarization camera Phoenix 5.0 MP PHX050S1-QC comprising a Sony IMX264MYR CMOS (Color) Polarsens sensor (LUCID Vision Labs, Inc., Richmond B.C, Canada) and a Universe Compact C-Mount 5MP 2/3'' 6mm f/2.0 lens (Universe, New York, USA).

As the amount of real data differs between objects, we follow common practice in instance-level object pose estimation literature by sampling around $15\% - 20\%$ of total data for training, and the rest for testing~\citep{gao2021polarimetric}, which results in 200-300 sets of real polarization images as training data for each object, and 1000-2000 sets of images as testing. 

We ensure that the poses for the data split of the rendered synthetic domain are similar to the poses of the real domain in terms of overall distribution, to ensure comparability when analyzing the domain shift and the influence of our proposed self-supervision scheme later. 
The predicted bounding box crops out the region containing the object of interest and is resized to $256 \times 256$ as inputs to the networks. The predicted object mask serves as input to the physical model to produce only object-related polarimetric parameters as well as shape priors.

\subsection{Implementation Details}
\label{subsec:implementation} 
We implement our model using Pytorch~\citep{paszke2019pytorch} and train on an NVIDIA 2080 GPU and using ADAM optimizer~\citep{kingma2014adam} on a commodity desktop PC with an Intel i7 CPU processor and 32GB RAM. The teacher and student networks are trained for 100 epochs for each object individually, both for synthetic and real data. The initial learning rate is set to $1 \times {10}^{-4}$, and halved every 25 epochs. The weights for all encoders are initialized with ImageNet weights. For synthetic pre-training, we use a batch size of 8, and for the self-supervised training on real data a batch size of 4.

\subsection{Refractive Indices}
As a material-related coefficient, the refractive index $\eta$ for each object is listed in Table~\ref{tab:refindex}. The index serves as input to both the forward and inverted physical model.
The refractive index is assumed to be known, but it has only a minor influence on object pose predictions (cf. also PPP-Net~\citep{gao2021polarimetric}). In terms of objects with different composite materials, we can observe the plastic cup in our experiments, where the plastic material is slightly different for the beige and green parts and the texture changes as well. Given the results for the different refractive indices (cf. Table A6. Refractive Index Ablation. in the Supp.Mat. of PPP-Net~\citep{gao2021polarimetric}), we expect that objects with different composite materials can still be handled. An extensive study of composite objects is out of scope for this work, as the PhoCal dataset~\citep{PhoCal} does not include other such objects.

\begin{table}[!b]
\centering
\caption{\textbf{Refractive Indices. }}
\begin{tabular}{l|c|c} 
    \shline
    Object & Material & Refractive Index \\ 
    \hline
    Fork & stainless steel & 2.75 \\ 
    Knife & stainless steel & 2.75 \\ 
    Bottle & glass & 1.52 \\ 
    Cup & plastics & 1.50 \\ 
    \shline
\end{tabular}

\label{tab:refindex}
\end{table}

\begin{table*}[!h]
\centering
\footnotesize
\caption{\textbf{\textbf{S}$^{2}$\textbf{P}$^{3}$ Quantitative Results. } Average recall of ADD(-S) metric is reported for different objects with increasing photometric complexity. Self6D++ from~\citep{wang2021occlusion}. PPP-Net from~\citep{gao2021polarimetric}.
}
\resizebox{0.8\textwidth}{!}{
\begin{tabular}{l|l|c|c|c|c|c}
\shline
\multicolumn{1}{l|}{Methods}  & \multicolumn{1}{l|}{Training}  & \multicolumn{1}{c}{Cup} & \multicolumn{1}{c}{Fork} & \multicolumn{1}{c}{Knife} & \multicolumn{1}{c|}{Bottle} & \multicolumn{1}{c}{Mean}  \\

\hline

PPP-Net & Supervised & 91.4 & 91.7 & 90.0 & 89.4 & 90.6 \\
\hline
Self6D++ & Self-Supervised (RGB-D) & 68.4 & 14.3 & 17.8 & 33.5 & 34.0 \\
\textbf{S}$^{2}$\textbf{P}$^{3}$ (Ours) & Self-Supervised (\textbf{RGB-P}) & \textbf{93.8} & \textbf{72.4} & \textbf{78.4} & \textbf{78.2} & \textbf{80.7} \\

\shline
\end{tabular}
}

\label{tab:full_model}
\end{table*}

\subsection{Evaluation Metrics. }
The results are evaluated using the common Average Distance of Distinguishable Model Points (ADD) metric~\citep{hinterstoisser2012model} for non-symmetrical objects, in which $10\%$ of the object’s diameter is set as the threshold to judge the average deviation of the transformed model points. For symmetric objects, the average deviation to the closest model points is measured as in the Distance of Indistinguishable Model Points (ADD-S) metric~\citep{hodavn2016evaluation}. 

\subsection{Quantitative Results - Baseline Comparisons}
\label{subsec:quantitative}

\textbf{S}$^{2}$\textbf{P}$^{3}$ proposes to leverage polarimetric information for self-supervised 6D object pose estimation and focuses on photometrically challenging objects, where self-supervised RGB-D methods may fail due to inherent sensor data artifacts, and supervised approaches, either RGB-only or RGB-P methods as e.g. PPP-Net~\citep{gao2021polarimetric}, would require a large amount of annotated real data. 
Therefore, the experiments are deliberately chosen to analyse the multi-modal self-supervision through the physical constraints, its loss functions, as well as the architecture and design choices for the student-teacher scheme in the ablation studies to yield best scientific insights into self-supervised polarimetric 6D pose estimation.  
As such, we compare \textbf{S}$^{2}$\textbf{P}$^{3}$ against PPP-Net~\citep{gao2021polarimetric} on our data split, as a very strong supervised baseline, in order to analyze the effect of self-supervision.
PPP-Net already outperforms other strong state-of-the-art RGB-only methods as reported in~\citep{gao2021polarimetric}, and is thus a valid upper threshold for comparison. 
Self6D++~\citep{wang2021occlusion} is the state-of-the-art self-supervised RGB-D method on many standard benchmark datasets, and is thus chosen for establishing polarimetric self-supervision in \textbf{S}$^{2}$\textbf{P}$^{3}$ as a strong baseline.
See also the qualitative results in Figures~\ref{fig:qualitative} and~\ref{fig:qualitative_zoom}, also including occlusions of the object as in Figures~\ref{fig:qualitative_occlusions} and~\ref{fig:qualitative_occlusions_zoom}, for visual results which are discussed later in more detail.

We prove the effectiveness of the self-supervision pipeline by quantitative results in Table~\ref{tab:full_model}. 
Please note, that PPP-Net (trained on annotated real data) is the identical network as we use in our teacher model but without the differentiable renderer. In our full model \textbf{S}$^{2}$\textbf{P}$^{3}$ however, we do not train the teacher in a supervised manner on real data, but only pre-train it on the synthetic data. Then, the weights of the teacher are frozen and it only provides weak pseudo-labels on real data for the teacher-student scheme.
Our model \textbf{S}$^{2}$\textbf{P}$^{3}$, consistently outperforms the self-supervised learning-based state-of-the-art RGB-D method Self6D++ by ~\citep{wang2021occlusion}
~\footnote{Self6D++ is trained and tested on our dataset, with RGB-D information from Realsense L515 sensor}, and even reaches comparable performance against the fully supervised upper bound baseline~\citep{gao2021polarimetric} for photometrically complex objects.

\subsection{Ablation Studies}
\label{subsec:ablation}

Our evaluation comprises several ablation studies to analyze the nuances of our model's components.
We assess performance variations between synthetic and real data domains, particularly in the absence of self-supervision, to answer the question: how well can the student and the teacher network perform on real data, when trained in a supervised fashion on synthetic or real data, respectively, and how much performance gain does \textbf{S}$^{2}$\textbf{P}$^{3}$ achieve when supervising on synthetic data only and performing self-supervision with real data.
We further explore the impact of the student's architecture within the student-teacher paradigm, focusing on whether a lightweight student could match or outperform the teacher when refined on real data. Or to put it simple: Do we need a large student model, identical to the teacher network with a decoder and dedicated geometrical predictions? Or is the design choice of \textbf{S}$^{2}$\textbf{P}$^{3}$ to directly regress the 6D pose for the student beneficial?
Additionally, we dissect the influence of individual loss components, emphasizing the significance of our physically-induced self-supervised loss.
And finally investigate the role of depth versus polarimetric information, gauging their relative contributions to the model's efficacy.

\paragraph{Ablation on Domain Shift -\\\textbf{S}$^{2}$\textbf{P}$^{3}$'s Self-Supervision}

\begin{table*}[!t]
\centering
\footnotesize
\caption{
\textbf{Domain Shift and \textbf{S}$^{2}$\textbf{P}$^{3}$'s Self-Supervision.} Average recall of ADD(-S) metric is reported for different objects with increasing photometric complexity for the student and teacher network individually, when trained in a supervised setting on either real or synthetic data and tested on real data. The full \textbf{S}$^{2}$\textbf{P}$^{3}$ pipeline, with synthetic pre-training, and self-supervised training of the student on non-annotated real data, is also reported for comparison.
"Teacher $\dagger$" as upper bound is identical to PPP-Net~\citep{gao2021polarimetric}.
"Student $\star$" corresponds to the setting of \textbf{S}$^{2}$\textbf{P}$^{3}$ \textbf{before} applying our proposed self-supervision scheme.
}
\resizebox{0.99\textwidth}{!}{
\begin{tabular}{l|l|c|c|c|c|c|c|c}
\shline
\multicolumn{1}{l|}{Configuration}  & \multicolumn{1}{c|}{Supervised} & \multicolumn{1}{c|}{Self-Supervised}  & \multicolumn{1}{c|}{Tested on}  & \multicolumn{1}{c}{Cup} & \multicolumn{1}{c}{Fork} & \multicolumn{1}{c}{Knife} & \multicolumn{1}{c|}{Bottle} & \multicolumn{1}{c}{Mean}  \\
\hline
$\text{Student}$ & \multicolumn{1}{c|}{Real} & - & Real & 86.4 & 88.0 & 91.1 & 80.4 & 86.5 \\
$\text{Teacher}$ $\dagger$ & \multicolumn{1}{c|}{Real} & - & Real & \textbf{91.4} & \textbf{91.7} & \textbf{90.0} & \textbf{89.4} & \textbf{90.6} \\

\hline\hline
$\text{Student}$ $\star$ & \multicolumn{1}{c|}{Synthetic} & - & Real & 53.7 & 64.4 & 46.1 & 47.5 & 52.9 \\
$\text{Teacher}$ & \multicolumn{1}{c|}{Synthetic} & - & Real & 72.3 & \textbf{75.0} & 67.3 & 76.2 & 72.7 \\

\hline
\textbf{S}$^{2}$\textbf{P}$^{3}$ (Ours) & \multicolumn{1}{c|}{Syn. (Pre-trained)} & \multicolumn{1}{c|}{Real} & \multicolumn{1}{c|}{Real} & \textbf{93.8} & 72.4 & \textbf{78.4} & \textbf{78.2} & \textbf{80.7} \\

\shline
\end{tabular}
}

\label{tab:abl:domain}
\end{table*}

Table~\ref{tab:abl:domain} summarizes the results when training the individual student and teacher network separately (not within the \textbf{S}$^{2}$\textbf{P}$^{3}$ training scheme), without the differentiable renderer, with supervision on the pose estimation as in the synthetic pre-training. We differentiate whether training is performed on annotated real or synthetic data, and test on real data. As expected, the student and teacher networks, perform worse on real data when trained on synthetic data only, due to the domain shift, compared to training on real data (compare top and lower rows of Table~\ref{tab:abl:domain} for student and teacher, respectively). The larger teacher network with a dedicated decoder and explicit intermediate geometrical representations, which is identical to PPP-Net~\citep{gao2021polarimetric} and marked with $\dagger$ in the table, outperforms the smaller student network when trained in a supervised fashion in both scenarios. Our full pipeline of \textbf{S}$^{2}$\textbf{P}$^{3}$(where the student is trained self-supervised on real data and the teacher weights are fixed), with our proposed small student network and a teacher, which are both only pre-trained on synthetic data (i.e., the synthetically pre-trained networks correspond to the numbers of the lower part of Table~\ref{tab:abl:domain}), achieves impressive results without being trained on annotations from real images due to our proposed self-supervision paradigm. \textbf{S}$^{2}$\textbf{P}$^{3}$ even partly outperforms the fully supervised training on real data (cf. top rows against \textbf{S}$^{2}$\textbf{P}$^{3}$) and achieves comparable results to PPP-Net as fully supervised upper boundary (indicated by $\dagger$). 
Notably, the self-supervision of \textbf{S}$^{2}$\textbf{P}$^{3}$ improves the results against the synthetically pre-trained student network (cf. Table~\ref{tab:abl:domain} "Student $\star$" against \textbf{S}$^{2}$\textbf{P}$^{3}$). While this trend holds true for all objects, the observation from before is not as significant for the fork, which may result from large occlusions for this object in the majority of the data (cf. Figures~\ref{fig:qualitative_occlusions} and~\ref{fig:qualitative_occlusions_zoom} where the fork is inside the cup).

\begin{table}[!b]
\centering
\footnotesize
\caption{\textbf{Ablation on Network Architecture. } We compare different architecture designs for the student network, i.e., our small student and a larger student which would be equivalent to the teacher architecture. Our proposed self-supervised student network (Ours) achieves best results across all objects. We report average recall of ADD(-S) metric.}
\resizebox{\columnwidth}{!}{
\begin{tabular}{l|l|c|c|c|c|c}
\shline
\multicolumn{1}{l|}{Config}  & \multicolumn{1}{l|}{Self-Sup.}  & \multicolumn{1}{c}{Cup} & \multicolumn{1}{c}{Fork} & \multicolumn{1}{c}{Knife} & \multicolumn{1}{c|}{Bottle} & \multicolumn{1}{c}{Mean}  \\
\hline
Our Student& None &	53.7&	64.4&	46.1&	47.5&	52.9\\
Large Student& None &	72.3&	75.0&	67.3&	76.2&	72.7\\
\hline
Our Student &\checkmark (\textbf{S}$^{2}$\textbf{P}$^{3}$) &	\textbf{93.8}&	\textbf{72.4}&	\textbf{78.4}&	\textbf{78.2}& 	\textbf{80.7}\\
Large Student&\checkmark&	88.6&	55.9&	69.4&	77.8&	73.0\\
\shline
\end{tabular}
}
\label{tab:ablation_architecture}
\end{table}

\begin{figure*}[!t]
      \centering
      \includegraphics[width=1.0\textwidth]{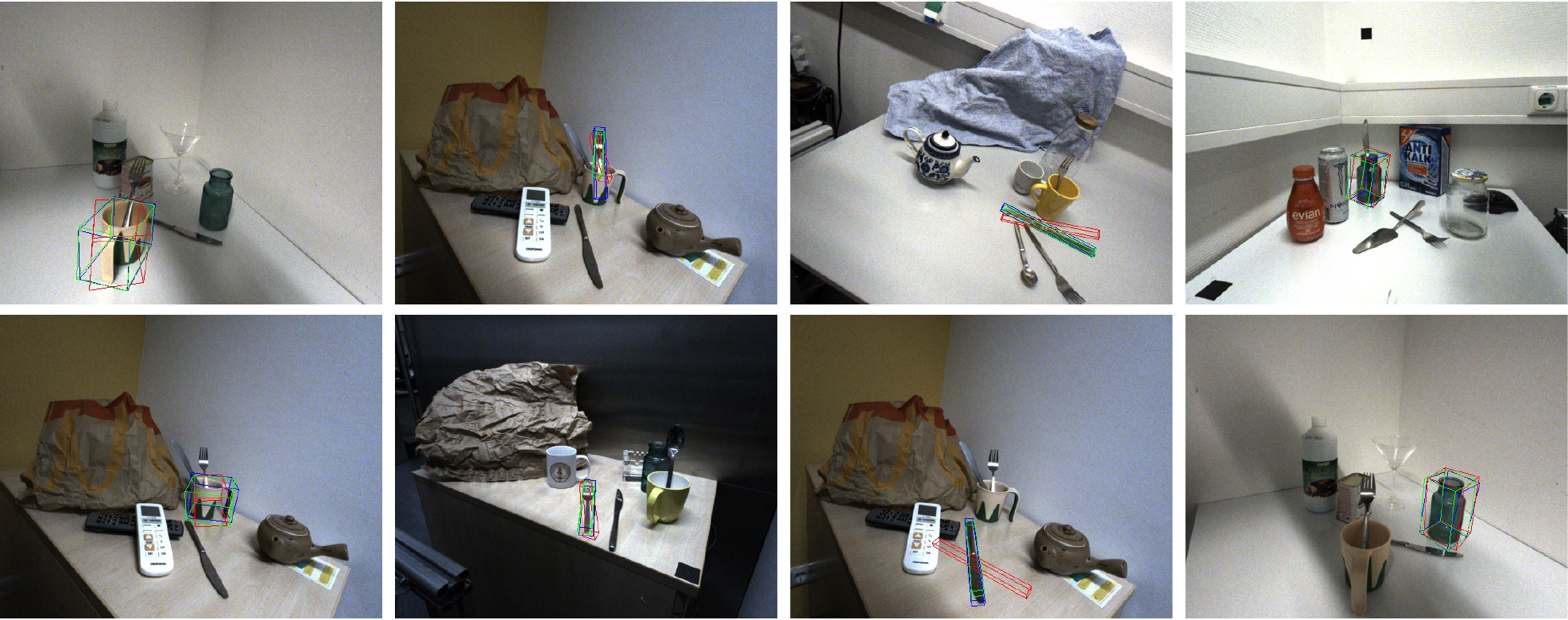}
      \caption{\textbf{\textbf{S}$^{2}$\textbf{P}$^{3}$ Qualitative Results Before and After Self-Supervision. } The projected bounding boxes in \textit{blue}, \textit{red} and \textit{green} represent the ground-truth 6D object poses, the results before and after applying self-supervision, respectively.}
      \label{fig:qualitative}
\end{figure*}

\begin{figure*}[!t]
      \centering
      \includegraphics[width=1.0\textwidth]{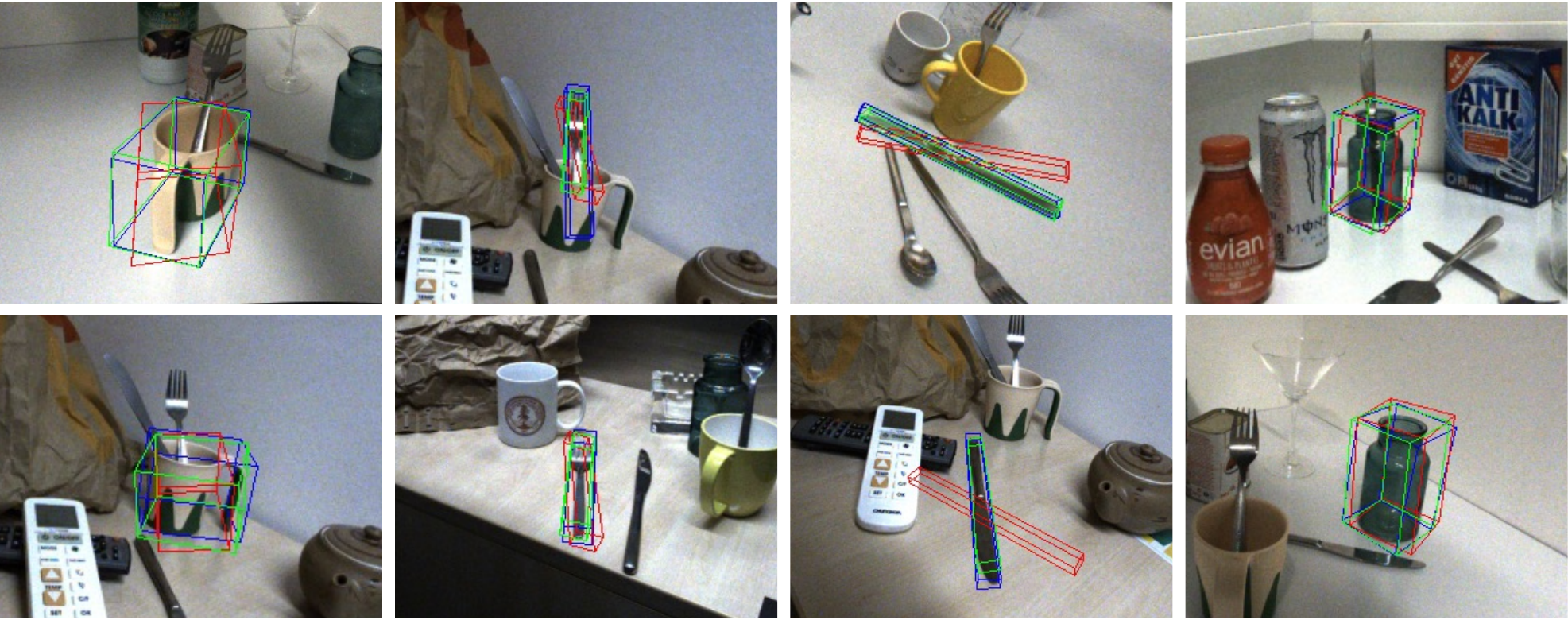}
      \caption{\textbf{\textbf{S}$^{2}$\textbf{P}$^{3}$ Qualitative Results Before and After Self-Supervision (zoomed-in from Figure~\ref{fig:qualitative}). } The projected bounding boxes in \textit{blue}, \textit{red} and \textit{green} represent the ground-truth 6D object poses, the results before and after applying self-supervision, respectively.}
      \label{fig:qualitative_zoom}
\end{figure*}

\begin{figure*}[!t]
      \centering
      \includegraphics[width=1.0\textwidth]{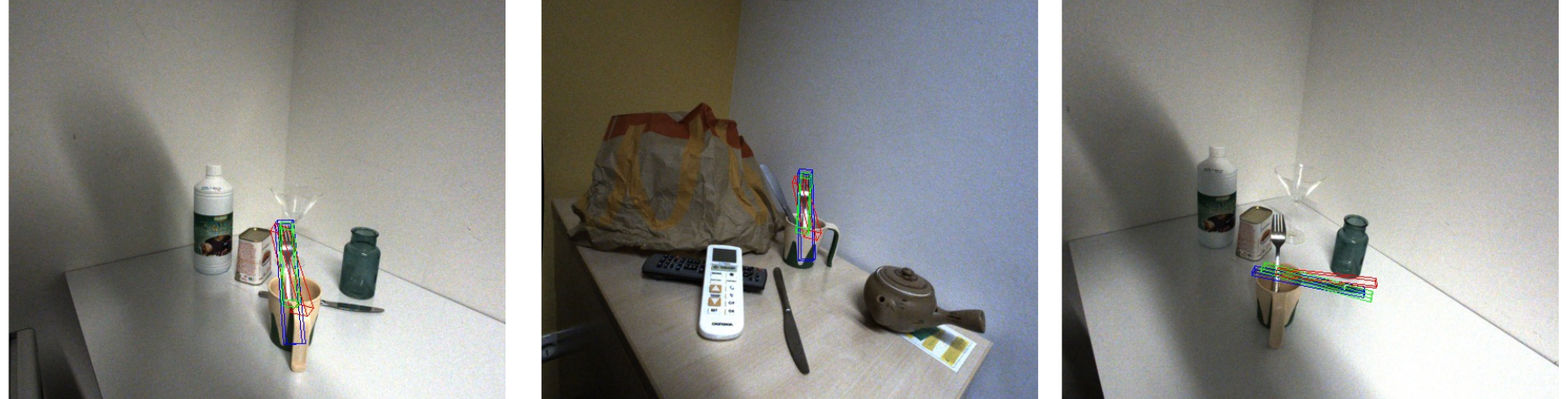}
      \caption{\textbf{\textbf{S}$^{2}$\textbf{P}$^{3}$ Qualitative Results Before and After Self-Supervision with Occlusions. } The projected bounding boxes in \textit{blue}, \textit{red} and \textit{green} represent the ground-truth 6D object poses, the results before and after applying self-supervision, respectively.}
      \label{fig:qualitative_occlusions}
\end{figure*}

\begin{figure*}[!t]
      \centering
      \includegraphics[width=1.0\textwidth]{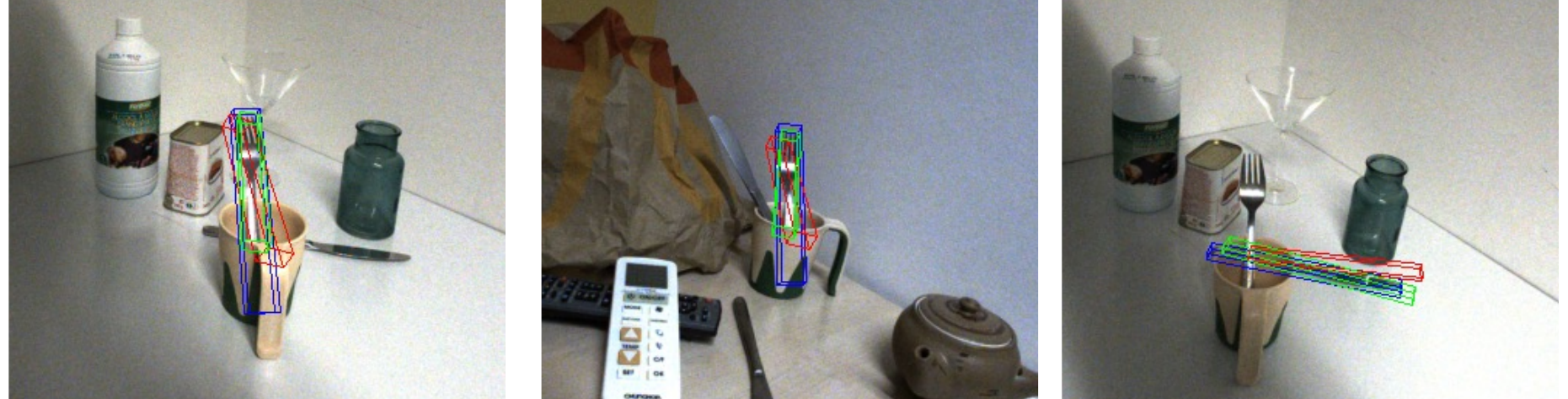}
      \caption{\textbf{\textbf{S}$^{2}$\textbf{P}$^{3}$ Qualitative Results Before and After Self-Supervision with Occlusions (zoomed-in from Figure~\ref{fig:qualitative_occlusions}). } The projected bounding boxes in \textit{blue}, \textit{red} and \textit{green} represent the ground-truth 6D object poses, the results before and after applying self-supervision, respectively.}
      \label{fig:qualitative_occlusions_zoom}
\end{figure*}

\paragraph{Ablation on Network Architecture - Exchanging the Student}
We follow the motivation to utilize a lightweight student for faster inference. We exchange the network architecture for the student network in \textbf{S}$^{2}$\textbf{P}$^{3}$, with the one that is normally used as teacher, i.e., instead of the network in Figure~\ref{fig:pipeline_student} we use the one of Figure~\ref{fig:pipeline_teacher} as the student, to analyze the influence of the larger network with its dedicated decoder and intermediate geometrical representations.
While intermediate geometrical outputs of the large student are beneficial during pre-training in a supervised fashion (cf. first two rows in Table~\ref{tab:ablation_architecture}), those outputs introduce more optimization objectives when used as the student network to learn the 6D pose of the object. The ablation in Table~\ref{tab:ablation_architecture} demonstrates that the lightweight student (Our Student) can achieve better performance than a larger student (Large Student) network after fine-tuning the student on real data with our student-teacher training scheme and self-supervision through physical constraints $\mathcal{L}_{physics}$. The additional parameters and the intermediate geometrical representations of the large student make convergence more difficult. Still, the physical constraints improve the large student significantly after self-supervision (cf. Large student with None and with Self-Supervision). The ablations demonstrate that the lightweight student can achieve better performance than the larger student network after fine-tuning on real data with our self-supervision scheme through physical constraints $\mathcal{L}_{physics}$, as employed in \textbf{S}$^{2}$\textbf{P}$^{3}$.

\paragraph{Ablation on Loss Terms}
We first verify the influence of various loss terms by training the network without each specific loss term for the self-supervision stage as summarized in Table~\ref{tab:ablation_loss}. We find that the direct geometrical point matching loss of $\mathcal{L}_{pose}$ is crucial to self-supervision. Without enforcing $\mathcal{L}_{pose}$ for the student against the weak pseudo-labels of the teacher, the training would easily diverge. 
The physically-induced self-supervised loss $\mathcal{L}_{physics}$, that is derived from our invertible physical derivations, indicates a larger impact on training results compared to geometrical supervision signals from the teacher network, e.g., $\mathcal{L}_{normal}$ and $\mathcal{L}_{mask}$. 
The captured real polarimetric images contain more robust underlying object shape information compared to the output of the differentiable renderer. 
The overall performance of the model reaches best accuracy metrics for all objects with varying photometric complexity when all loss ingredients are present, as indicated in the last row of Table~\ref{tab:ablation_loss}.

These results indicate, that the convergence of the student can only be guaranteed when weak labels of the teacher network roughly guide the pose predictions. 
One reason to explain such behavior, is that the differentiable renderer would be completely unconstrained without $\mathcal{L}_{pose}$, thus potentially rendering outputs with pose predictions that are out of the field of view.
Dense supervision of the appearance and geometric representations after differentiable rendering further improve the networks performance, while the boost in pose accuracy is most noticeable with our proposed self-supervised physically-induced loss formulation. The contribution of the self-supervision is also apparent in the qualitative results in Figures~\ref{fig:qualitative} and~\ref{fig:qualitative_zoom}. The projected bounding boxes in $\emph{green}$ show better alignment with ground truth ($\emph{blue}$) after self-supervision, compared against predictions of the pre-trained teacher ($\emph{red}$). Figures~\ref{fig:qualitative_occlusions} and~\ref{fig:qualitative_occlusions_zoom} show additional results for cases where part of the object, here \textit{fork} and \textit{knife}, is occluded.

\begin{table}[!t]
\centering
\footnotesize
\caption{\textbf{Ablation on Loss Terms. }Average recall of ADD(-S) metric is reported.}

\resizebox{\columnwidth}{!}{
\begin{tabular}{l|c|c|c|c|c}
\shline
\multicolumn{1}{c|}{Methods}  & \multicolumn{1}{c}{Cup} & \multicolumn{1}{c}{Fork} & \multicolumn{1}{c}{Knife} & \multicolumn{1}{c|}{Bottle} & \multicolumn{1}{c}{Mean}  \\

\hline
w/o $\mathcal{L}_{pose}$ & 6.8 & 0.2 & 2.3 & 0.6 & 2.5 \\
w/o $\mathcal{L}_{physics}$ & 71.8 & 72.1 & 70.8 & 74.4 & 72.3 \\
w/o $\mathcal{L}_{normal}$ & 87.5 & 61.0 & 67.3 & 74.9 & 72.7 \\
w/o $\mathcal{L}_{mask}$ & 89.9 & 64.9 & 70.1 & 72.7 & 74.4 \\
\hline
\textbf{S}$^{2}$\textbf{P}$^{3}$ (Ours) & \textbf{93.8} & \textbf{72.4} & \textbf{78.4} & \textbf{78.2} & \textbf{80.7} \\
\shline
\end{tabular}
}

\label{tab:ablation_loss}
\end{table}

\begin{table}[!b]
\centering
\footnotesize
\caption{\textbf{\textbf{S}$^{2}$\textbf{P}$^{3}$ Ablations on Depth Modality. } Average recall of ADD(-S) metric is reported.}
\footnotesize
\resizebox{\columnwidth}{!}{
\begin{tabular}{l|c|c|c|c|c}
\shline
\multicolumn{1}{c|}{Ours with:}  & \multicolumn{1}{c}{Cup} & \multicolumn{1}{c}{Fork} & \multicolumn{1}{c}{Knife} & \multicolumn{1}{c|}{Bottle} & \multicolumn{1}{c}{Mean}  \\

\hline
RGB-D Chamfer & \textbf{100.0} & 11.6 & 59.1 & 40.7 & 52.9 \\
RGB-D Pixel-wise&	86.8&	32.3&	62.5&	50.3&	58.0\\
\hline
RGB-P (\textbf{S}$^{2}$\textbf{P}$^{3}$) & 93.8 & \textbf{72.4} & \textbf{78.4} & \textbf{78.2} & \textbf{80.7} \\
\shline
\end{tabular}
}

\label{tab:ablation_depth}
\end{table}

\paragraph{Ablation on Modalities}

\noindent\textit{RGB-Texture Supervision. }
For textureless and transparent objects, the rendered object texture will only be white, since it does not have any color (cf. also Figure 7 in PhoCal~\citep{PhoCal} and Figure 5 in PPP-Net~\citep{gao2021polarimetric}). This would reduce the RGB-texture loss essentially to the mask loss in our pipeline. Hence, we eliminate the need for texture rendering and instead rely on the physical properties of polarized light.\\

\noindent\textit{Depth Supervision. }
To analyze the importance of accurate and reliable geometric representations for the task of 6D object pose estimation, we train our pipeline with depth maps from a direct time of flight (D-ToF) sensor and compare it against the polarimetric \textbf{S}$^{2}$\textbf{P}$^{3}$ method with our physically-induced self-supervised loss.
For this purpose, we adapt our network to have an additional loss term utilizing depth information aside from having almost all other components unchanged. 
We let the differentiable renderer of the student network additionally render depth maps $\mathbf{D}^R$ given the predicted pose $\mathbf{\hat{P}_s}$, and employ a chamfer distance loss $\mathcal{L}_{chamfer}$ between the point cloud $\mathbf{P}^R$ back-projected from the rendered depth $\mathbf{D}^R$ and the point cloud $\mathbf{P}$ back-projected from the depth map in the polarization camera coordinate system, to optimize for alignments without explicit 3D-3D correspondence registrations as:

\begin{equation} 
\label{eq:loss_chamfer}
\begin{split}
    \mathcal{L}_{chamfer} & = \underset{\mathbf{p} \in \mathbf{P}}{\avg} \underset{\mathbf{p}^r \in \mathbf{P}^R}{\min} \| \mathbf{p} - \mathbf{p}^r \|_2 \\
    & + \underset{\mathbf{p}^r \in \mathbf{P}^R}{\avg} \underset{\mathbf{p} \in \mathbf{P}}{\min} \| \mathbf{p} - \mathbf{p}^r \|_2 .
\end{split}
\end{equation}

Besides adding $\mathcal{L}_{chamfer}$ to the pipeline, we remove the $\mathcal{L}_{physics}$ to have a fair comparison of the effectiveness of direct spatial cues from depth and object shape cues from polarimetric physical properties. The results listed in Table~\ref{tab:ablation_depth} indicate the depth cues can be beneficial when the quality is reliable, i.e., the performance on the \textit{cup} peeks when $\mathcal{L}_{chamfer}$ is introduced to the pipeline.

\begin{figure*}[!t]
      \centering
      \includegraphics[width=\textwidth]{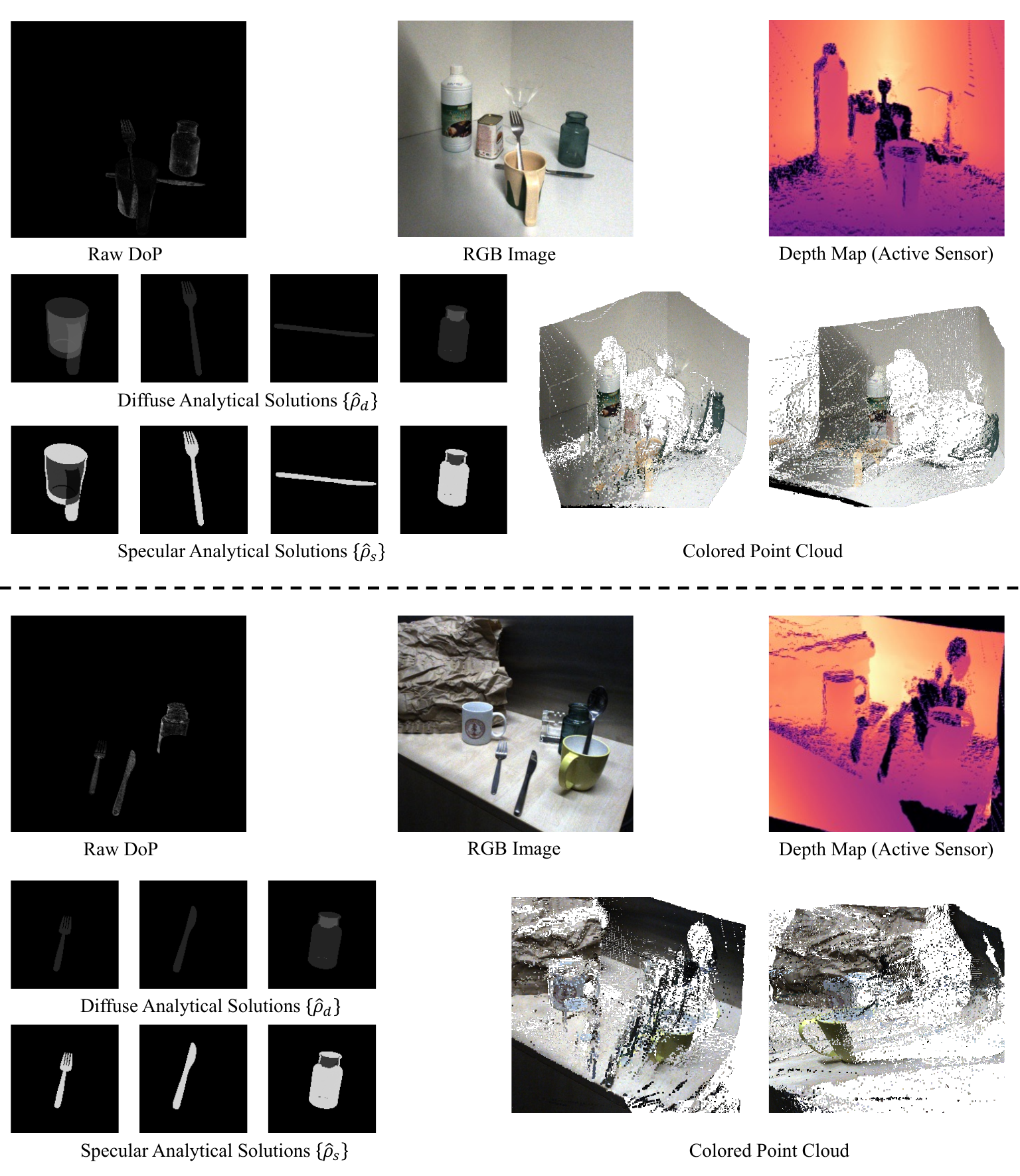}
      \caption{\textbf{Examples of Polarimetric and Depth Quality. }
      }
      \label{fig:dopvsdepth}
\end{figure*}

We conduct additional ablations using a pixel-wise depth loss instead of the chamfer distance loss, as reported in Table~\ref{tab:ablation_depth}. The experiment illustrates that also with the pixel-wise depth loss, inaccurate depth information would inject incorrect geometric guidance into the pipeline, leading to degraded performance on photometrically challenging objects.

The inherent limitations of the depth sensor cause severe degradation of the depth quality~\citep{jung2022my}.
The reflective and semi-transparent objects are measured incorrectly due to reflective and translucent object materials.
This is also illustrated in detailed large figures of real examples in Figure~\ref{fig:dopvsdepth}.
In such cases, the strong signal coming from depth alignment loss introduces incorrect spatial awareness, leading to low pose prediction performance. 

On the contrary, the shape of the object that is encoded in the polarimetric image modality can provide stable geometric information for objects of all material characteristics presented here, across a variety of photometric complexity, e.g., from a matte plastic cup, to reflective stainless steel cutlery, and translucent and transparent colored glass objects.
The analytically retrieved diffuse and specular solutions after the differentiable renderer are stable across all discussed objects. These polarization properties are computed through our invertible model and then utilized in the physics-induced self-supervision scheme against the raw DoP illustrated on the top left in Figure~\ref{fig:dopvsdepth}. Please note that $\mathcal{L}_{physics}$ is a pixelwise minimum loss of the diffuse and specular reflection.

\paragraph{Runtime Analysis}
On a desktop PC with an Intel i7 4.20GHz CPU and an NVIDIA 2080 GPU, given a $512 \times 612$ image, our student network takes $\approx 7.3$ ms for inferring the 6D pose for a single object, which is around $30\%$ faster than the teacher model. Additionally, the preprocessing for the physical prior calculation takes 13.0 ms, and the object detection takes 15.4 ms.

\section{Conclusion}

\label{conclusion}
\paragraph{Limitations}
The performed experiments highlight the importance of reliable geometric priors for the task of 6D object pose estimation. When the quality of the depth map is reliable and accurate, the spatial loss term introduced by the source depth map may lead to better performance than pure object-shape-based optimization through polarization. The current model focuses on instance-level pose estimation and does not generalize to unseen objects during training. An interesting future direction is to include the idea in a category-level pipeline.

\paragraph{Self-Supervised Polarimetric Pose Prediction}
This paper bridges two worlds and combines a hybrid model for polarimetric pose estimation that fuses an invertible physical model with neural shape extraction from data within a self-supervised framework. \textbf{S}$^{2}$\textbf{P}$^{3}$ solves instance-level object pose estimation from polarimetric images without annotated real data. In our proposed pipeline, a teacher pre-trained on a small set of synthetic renderings ensures convergence of a lightweight student network through weak pseudo-labels. Our employed differentiable renderer additionally provides the appearance and geometric outputs and enables self-supervision. 
\textbf{S}$^{2}$\textbf{P}$^{3}$ outperforms methods that use depth measurements from active sensors for photometrically challenging objects. We achieve this by carefully integrating distinct design choices in the student-teacher architecture and proposing our invertible physical model for self-supervision by leveraging XoP properties, instead of raw polarimetric data as in~\citep{CroMo}, to reduce the domain gap. Our contributions are validated through extensive ablation studies. 

Our experimental results show the importance of self-supervision through geometric and physical cues for the task of 6D pose estimation and yield scientific insights into the robustness of polarimetric images. Such observations are most noticeable for photometrically challenging, texture-less, reflective, or translucent objects.

\clearpage
\newpage

\backmatter

\bibliography{sn-bibliography}


\end{document}